\documentclass{article}
\usepackage{hello}

\usepackage[margin=1in]{geometry}

\usepackage{graphicx}
\usepackage{booktabs}
\usepackage{array}
\usepackage{amsmath}
\usepackage{amssymb}
\usepackage{amsfonts}
\usepackage{multirow}
\usepackage{verbatim}
\usepackage{caption}
\usepackage{longtable}
\usepackage{supertabular}

\usepackage{enumitem}
\usepackage{tablefootnote}
\usepackage[round,semicolon]{natbib}
\usepackage{colortbl}
\usepackage{xspace}
\usepackage{textcomp}
\usepackage{makecell}
\usepackage{multirow}
\usepackage{lscape} 
\usepackage{siunitx}

\usepackage{enumitem}
\usepackage{tablefootnote}

\usepackage{colortbl}
\usepackage{xspace}
\usepackage{textcomp}
\usepackage{makecell}
\usepackage{multirow}
\usepackage{lscape} 
\usepackage{siunitx}
\definecolor{dt}{gray}{0.7}
\usepackage{pifont}       % \ding{xx}
\usepackage{bbding}       % \Checkmark,\XSolid,... (需要和pifont宏包共同使用)
\usepackage{fontawesome}

\usepackage{amssymb}% http://ctan.org/pkg/amssymb
\usepackage{pifont}% http://ctan.org/pkg/pifont
\usepackage{scrextend}

\usepackage{array}
\usepackage{tgpagella}
\usepackage{latexsym}
\usepackage[T1]{fontenc}
\usepackage[utf8]{inputenc}
\usepackage{microtype}
\definecolor{mydarkblue}{rgb}{0,0.08,0.45}
\usepackage[colorlinks,citecolor=mydarkblue,urlcolor=mydarkblue,linkcolor=mydarkblue]{hyperref}
\usepackage{url}            % simple URL typesetting
\usepackage{nicefrac}       % compact symbols for 1/2, etc.
\usepackage{changepage}
\usepackage{xargs}          % Use more than one optional parameter in a new commands
\usepackage{wrapfig,lipsum,booktabs}
\usepackage{longtable}
\usepackage{subcaption}
\usepackage{endnotes}

\usepackage{pgfplots}
\usetikzlibrary{pgfplots.groupplots}
\pgfplotsset{compat=1.3}
\usepackage{tikz}
\usetikzlibrary{patterns}

\usepackage[most]{tcolorbox}

\usepackage[capitalize,noabbrev]{cleveref}
\crefname{section}{Section}{\S\S}
\Crefname{section}{Section}{\S\S}
\crefname{table}{Table}{Tables}
\crefname{figure}{Figure}{Figures}
\crefname{algorithm}{Algorithm}{}
\crefname{equation}{eq.}{}
\crefname{appendix}{Appendix}{}
\crefformat{section}{Section #2#1#3}
\usepackage{multicol}
\usepackage{multirow}

\usepackage{titlesec}
\titleformat*{\section}{\large\bfseries}

\definecolor{battleshipgrey}{rgb}{0.3, 0.3, 0.3}
\definecolor{brilliantrose}{rgb}{1.0, 0.33, 0.64}
\definecolor{americanrose}{rgb}{1.0, 0.01, 0.24}
\definecolor{jweigreen}{rgb}{0,0.45,0.24}
\definecolor{bluegray}{rgb}{0.1, 0.1, 0.4}
\definecolor{ao(english)}{rgb}{0.0, 0.5, 0.0}
\definecolor{blanchedalmond}{rgb}{1.0, 0.92, 0.8}
\definecolor{atomictangerine}{rgb}{1.0, 0.6, 0.4}
\definecolor{chocolate(web)}{rgb}{0.82, 0.41, 0.12}
\definecolor{bananayellow}{rgb}{1.0, 0.88, 0.21}
\definecolor{goldenbrown}{rgb}{0.6, 0.4, 0.08}
\definecolor{aliceblue}{rgb}{0.94, 0.97, 1.0}
\definecolor{beige}{rgb}{0.96, 0.96, 0.86}
\definecolor{babyblue}{rgb}{0.54, 0.81, 0.94}
\definecolor{camel}{rgb}{0.76, 0.6, 0.42}
\definecolor{cinnamon}{rgb}{0.82, 0.41, 0.12}
\definecolor{deepskyblue}{rgb}{0.0, 0.75, 1.0}
\definecolor{frenchblue}{rgb}{0.0, 0.45, 0.73}
\definecolor{classicrose}{rgb}{0.98, 0.8, 0.91}
\definecolor{frenchrose}{rgb}{0.96, 0.29, 0.54}
\definecolor{frenchlilac}{rgb}{0.53, 0.38, 0.56}
\definecolor{frenchbeige}{rgb}{0.65, 0.48, 0.36}
\definecolor{verylightgreen}{RGB}{240, 255, 235}
\definecolor{verylightred}{RGB}{255, 235, 235}
\definecolor{verylightyellow}{RGB}{255, 254, 235}
\definecolor{dt}{gray}{0.7}

\definecolor{forestgreen}{HTML}{2e7d43}
\definecolor{color1}{HTML}{FF9999}
\definecolor{color2}{HTML}{FF6666}
\definecolor{color3}{HTML}{FF3333}
\definecolor{color4}{HTML}{E60000}
\definecolor{color5}{HTML}{B30000}
\definecolor{color6}{HTML}{8CD98C}
\definecolor{color7}{HTML}{53c653}
\definecolor{color8}{HTML}{39ac39}
\definecolor{color9}{HTML}{2d862d}
\definecolor{color10}{HTML}{206020}
\definecolor{color11}{HTML}{cca300}

\usepackage{minitoc}
\usepackage{float}
\newlength\savewidth
\newcommand{\tablestyle}[2]{\setlength{\tabcolsep}{#1}\renewcommand{\arraystretch}{#2}\centering\footnotesize}

\newcommand{\scripttablestyle}[2]{\setlength{\tabcolsep}{#1}\renewcommand{\arraystretch}{#2}\centering\scriptsize}

\usepackage{amsmath,amsfonts,bm}

\def\eqref#1{equation~\ref{#1}}
\def\1{\bm{1}}

\DeclareMathAlphabet{\mathsfit}{\encodingdefault}{\sfdefault}{m}{sl}
\SetMathAlphabet{\mathsfit}{bold}{\encodingdefault}{\sfdefault}{bx}{n}

\title{
\textbf{
Qwen-VL: A Versatile Vision-Language Model for Understanding, Localization, Text Reading, and Beyond}
}

\author{
\large{}
Jinze Bai$^*$ \hspace{6mm} Shuai Bai$^*$ \hspace{6mm} Shusheng Yang$^*$ \hspace{6mm} Shijie Wang \hspace{6mm} Sinan Tan\\
Peng Wang \hspace{6mm} Junyang Lin \hspace{6mm} Chang Zhou$^{\dag}$ \hspace{6mm} Jingren Zhou
\\
\large{}
Alibaba Group
\\
\small{}
Code \& Demo \& Models: \ \ \url{https://github.com/QwenLM/Qwen-VL}
}

\date{}

\begin{document}

\doparttoc % Tell to minitoc to generate a toc for the parts
\faketableofcontents % Run a fake tableofcontents command for the partocs

\maketitle

\begin{abstract}
\noindent
In this work, we introduce the Qwen-VL series, a set of large-scale vision-language models (LVLMs) designed to perceive and understand both texts and images.
Starting from the Qwen-LM as a foundation, we endow it with visual capacity by the meticulously designed \text{(i) visual receptor}, \text{(ii) input-output interface}, \text{(iii) 3-stage training pipeline}, and \text{(iv) multilingual multimodal cleaned corpus}.
Beyond the conventional image description and question-answering, we implement the grounding and text-reading ability of Qwen-VLs by aligning image-caption-box tuples.
The resulting models, including Qwen-VL and Qwen-VL-Chat, set new records for generalist models under similar model scales on a broad range of visual-centric benchmarks (\emph{e.g.}, image captioning, question answering, visual grounding) and different settings (\emph{e.g.}, zero-shot, few-shot).
Moreover, on real-world dialog benchmarks, our instruction-tuned Qwen-VL-Chat also demonstrates superiority compared to existing vision-language chatbots.
All models are public to facilitate future research.
% Code, demo and models are available at \url{https://github.com/QwenLM/Qwen-VL}.
\end{abstract}

{\let\thefootnote\relax\footnotetext{$^*$Equal contribution, $^\dag$Corresponding author}}

% keywords can be removed
%\keywords{First keyword \and Second keyword \and More}

\begin{figure*}[h]
\centering
\includegraphics[width= 8cm]{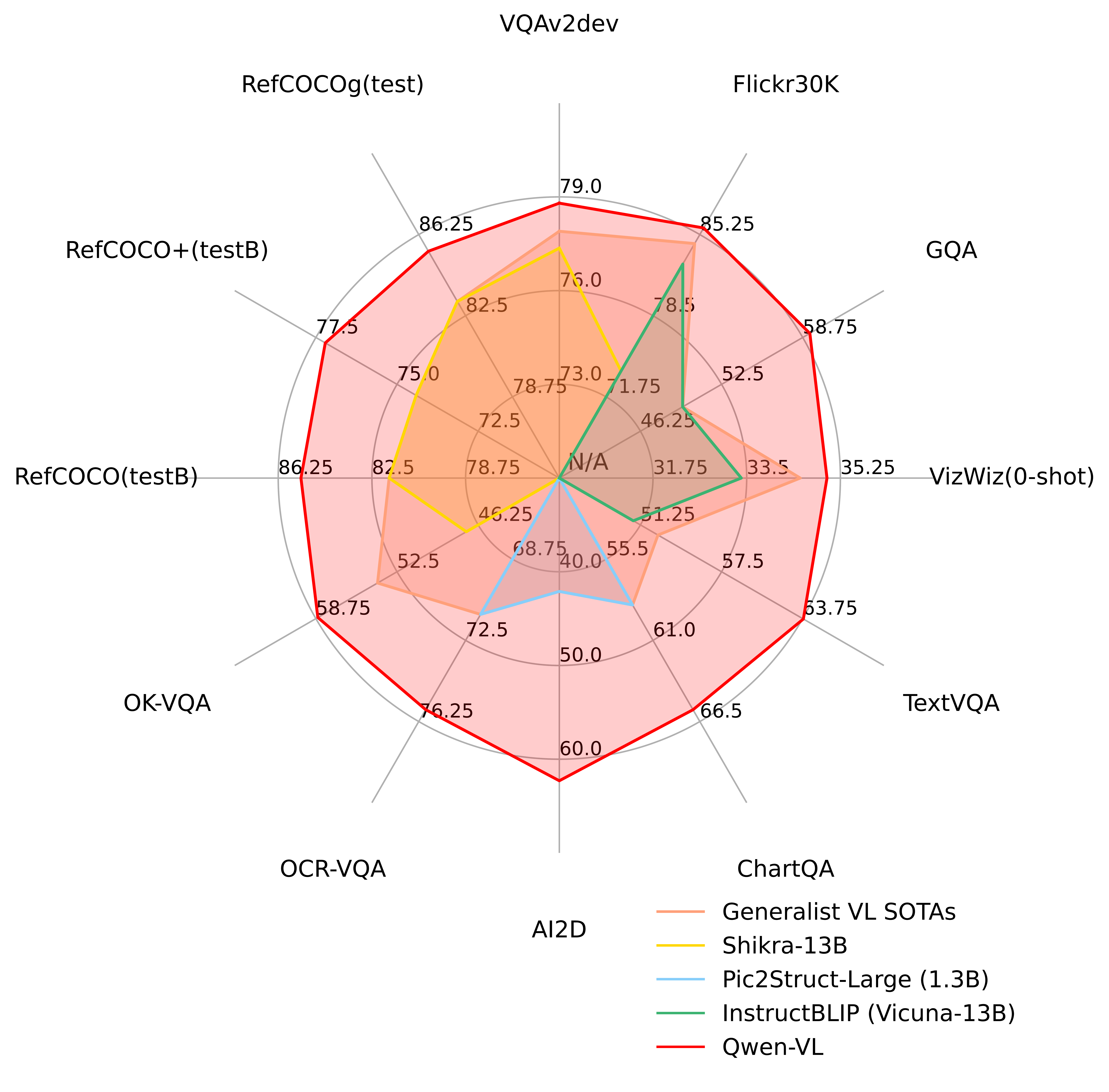}
   \caption{Qwen-VL achieves state-of-the-art performance on a broad range of tasks compared with other generalist models.}
\label{radar}
\end{figure*}

\begin{figure*}[t]
\centering
\includegraphics[width= 1\textwidth]{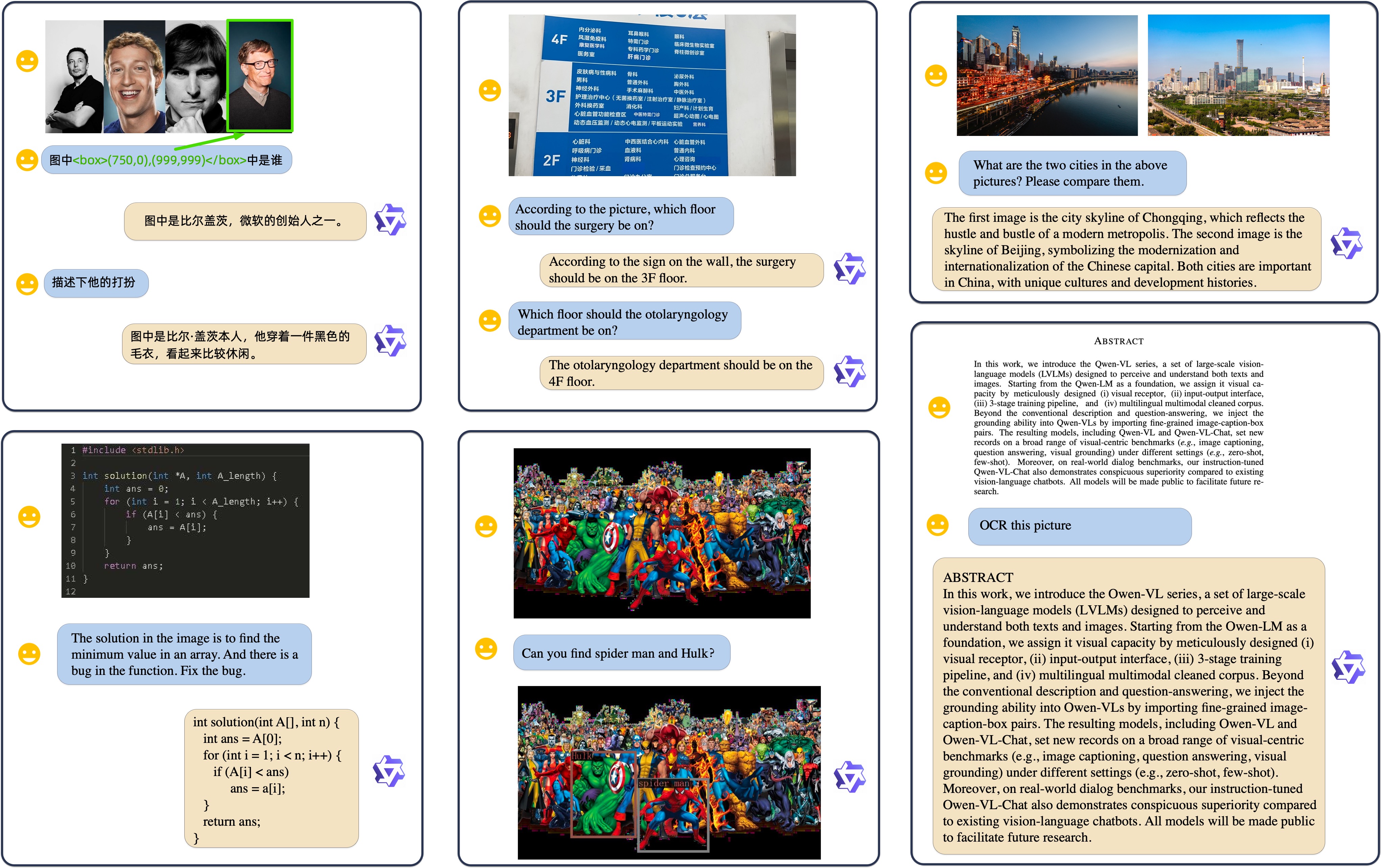}
\caption{Some qualitative examples generated by our Qwen-VL-Chat. Qwen-VL-Chat supports multiple image inputs, multi-round dialogue, multilingual conversation, text-reading, localization, fine-grained recognition and understanding ability.}
\label{example}
\end{figure*}

\section{Introduction}

Recently, Large Language Models (LLMs)~\citep{gpt3,gpt4,anil2023palm,gao2023llama,qwen7b} have attracted wide attention due to their powerful capabilities in text generation and comprehension. These models can be further aligned with user intent through fine-tuning instructions, showcasing strong interactive capabilities and the potential to enhance productivity as intelligent assistants.
However, native large language models only live in the pure-text world, lacking the ability to handle other common modalities (such as images, speech, and videos), resulting in great restrictions on their application scope.
Motivated by this, a group of Large Vision Language Models (LVLMs)~\citep{alayrac2022flamingo,chen2022pali, blip2,dai2023instructblip,kosmos,kosmos2,zhu2023minigpt,liu2023visual,ye2023mplug,mPLUG-DocOwl,shikra,li2023otter,videollama,emu,gpt4} have been developed to enhance large language models with the ability to perceive and understand visual signals. These large-scale vision-language models demonstrate promising potential in solving real-world vision-central problems.

Nevertheless, despite that lots of works have been conducted to explore the limitation and potency of LVLMs, current open-source LVLMs always suffer from inadequate training and optimization, thus lag far behind the proprietary models \citep{chen2022pali, chen2023pali, gpt4}, which hinders further exploration and application of LVLMs in open-source community.
What's more, as real-world visual scenarios are quite complicated, fine-grained visual understanding plays a crucial role for LVLMs to assist people effectively and precisely.
But only a few attempts had been made toward this direction \citep{kosmos2, shikra}, the majority of open-source LVLMs remain perceiving the image in a coarse-grained approach and lacking the ability to execute fine-grained perception such as object grounding or text reading.

In this paper, we explore a way out and present the newest members of the open-sourced Qwen families: Qwen-VL series.
Qwen-VLs are a series of highly performant and versatile vision-language foundation models based on Qwen-7B \citep{qwen7b} language model.
We empower the LLM basement with visual capacity by introducing a new visual receptor including a language-aligned visual encoder and a position-aware adapter.
The overall model architecture as well as the input-output interface are quite concise and we elaboratedly design a 3-stage training pipeline to optimize the whole model upon a vast collection of image-text corpus.

Our pre-trained checkpoint, termed Qwen-VL, is capable of perceiving and understanding visual inputs, generating desired responses according to given prompts, and accomplishing various vision-language tasks such as image captioning, question answering, text-oriented question answering, and visual grounding.
Qwen-VL-Chat is the instruction-tuned vision-language chatbot based on Qwen-VL.
As shown in Fig.~\ref{example}, Qwen-VL-Chat is able to interact with users and perceive the input images following the intention of users.

Specifically, the features of the Qwen-VL series models include:
\begin{itemize}

\item Leading performance: Qwen-VLs achieve top-tier accuracy on a vast of vision-centric understanding benchmarks compared to counterparts with similar scales. Besides, Qwen-VL's stuning performance covers not only the conventional benchmarks \emph{e.g.}, captioning, question-answering, grounding), but also some recently introduced dialogue benchmarks.

\item Multi-lingual: Similar to Qwen-LM, Qwen-VLs are trained upon multilingual image-text data with a considerable amount of corpus being in English and Chinese. In this way, Qwen-VLs naturally support English, Chinese, and multilingual instructions.

\item Multi-image: In the training phase, we allow arbitrary interleaved image-text data as Qwen-VL's inputs. This feature allows our Qwen-Chat-VL to compare, understand, and analyze the context when multiple images are given.

\item Fine-grained visual understanding: Thanks to the higher-resolution input size and fine-grained corpus we used in training, Qwen-VLs exhibit highly competitive fine-grained visual understanding ability. Compared to existing vision-language generalists, our Qwen-VLs possess much better grounding, text-reading, text-oriented question answering, and fine-grained dialog performance.
\end{itemize}

\section{Methodology}

\subsection{Model Architecture}
The overall network architecture of Qwen-VL consists of three components and the details of model parameters are shown in Table~\ref{tab:model_parameter}:

\textbf{Large Language Model}: Qwen-VL adopts a large language model as its foundation component. The model is initialized with pre-trained weights from Qwen-7B~\citep{qwen7b}.

\textbf{Visual Encoder}: The visual encoder of Qwen-VL uses the Vision Transformer (ViT)~\citep{dosovitskiy2020vit} architecture, initialized with pre-trained weights from Openclip's ViT-bigG~\citep{openclip}. During both training and inference, input images are resized to a specific resolution. The visual encoder processes images by splitting them into patches with a stride of 14, generating a set of image features.

\textbf{Position-aware Vision-Language Adapter}: To alleviate the efficiency issues arising from long image feature sequences, Qwen-VL introduces a vision-language adapter that compresses the image features. This adapter comprises a single-layer cross-attention module initialized randomly. The module uses a group of trainable vectors (Embeddings) as query vectors and the image features from the visual encoder as keys for cross-attention operations. This mechanism compresses the visual feature sequence to a fixed length of 256. The ablation about the number of queries is shown in Appendix \ref{app:n_queries}. Additionally, considering the significance of positional information for fine-grained image comprehension, 2D absolute positional encodings are incorporated into the cross-attention mechanism's query-key pairs to mitigate the potential loss of positional details during compression. The compressed image feature sequence of length 256 is subsequently fed into the large language model.

\begin{table}[h]
    \centering
    \caption{Details of Qwen-VL model parameters.}
    \begin{tabular}{cccc}
         \toprule
         Vision Encoder & VL Adapter & LLM & Total  \\
         \midrule
         1.9B & 0.08B & 7.7B & 9.6B \\
         \bottomrule
    \end{tabular}
    \label{tab:model_parameter}
\end{table}

\begin{figure*}[ht]
\centering
\includegraphics[width= 1\textwidth]{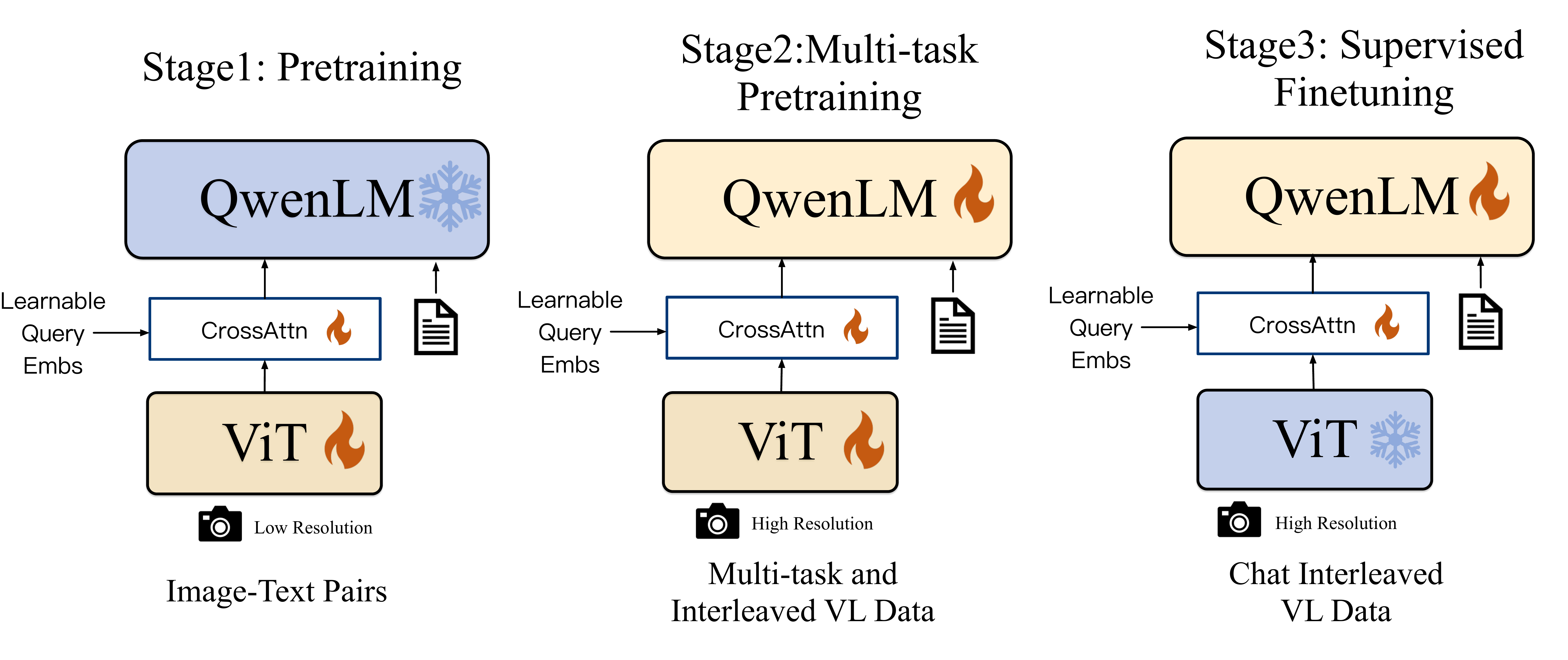}
    \caption{The training pipeline of the Qwen-VL series.}
\label{vl_train}
\end{figure*}

\subsection{Inputs and Outputs}
\textbf{Image Input}: Images are processed through the visual encoder and adapter, yielding fixed-length sequences of image features. To differentiate between image feature input and text feature input, two special tokens ($<$img$>$ and $<$/img$>$) are appended to the beginning and end of the image feature sequence respectively, signifying the start and end of image content. 

\textbf{Bounding Box Input and Output}: To enhance the model's capacity for fine-grained visual understanding and grounding, Qwen-VL's training involves data in the form of region descriptions, questions, and detections. Differing from conventional tasks involving image-text descriptions or questions, this task necessitates the model's accurate understanding and generation of region descriptions in a designated format. For any given bounding box, a normalization process is applied (within the range [0, 1000)) and transformed into a specified string format: "$(X_{top left}, Y_{top left}), (X_{bottom right}, Y_{bottom right})$". The string is tokenized as text and does not require an additional positional vocabulary. To distinguish between detection strings and regular text strings, two special tokens ($<$box$>$ and $<$/box$>$ are added at the beginning and end of the bounding box string. Additionally, to appropriately associate bounding boxes with their corresponding descriptive words or sentences, another set of special tokens ($<$ref$>$ and $<$/ref$>$) is introduced, marking the content referred to by the bounding box.

\section{Training}
As illustrated in Fig.~\ref{vl_train}, the training process of the Qwen-VL model consists of three stages: two stages of pre-training and a final stage of instruction fine-tuning training.

\subsection{Pre-training}

In the first stage of pre-training, we mainly utilize a large-scale, weakly labeled, web-crawled set of image-text pairs. Our pre-training dataset is composed of several publicly accessible sources and some in-house data. We made an effort to clean the dataset of certain patterns. As summarized in Table~\ref{tab:pretraining_data}, the original dataset contains a total of 5 billion image-text pairs, and after cleaning, 1.4 billion data remain, with 77.3\% English (text) data and 22.7\% Chinese (text) data.

\begin{table}[ht]
    \centering
    \caption{Details of Qwen-VL pre-training data. LAION-en and LAION-zh are the English 
 and Chinese language subset of LAION-5B~\citep{laion5b}. LAION-COCO~\citep{laioncoco} is a synthetic dataset generated from LAION-en. DataComp~\citep{datacomp} and Coyo~\citep{coyo} are collections of image-text pairs. CC12M~\citep{cc12m}, CC3M~\citep{cc3m}, SBU~\citep{sbu} and COCO Caption~\citep{cococaption} are academic caption datasets.} 
    \tablestyle{6pt}{1.1}
    \begin{tabular}{ll ccc}
         \toprule
         \textbf{Language} & \textbf{Dataset} & \textbf{Original} & \textbf{Cleaned} & \textbf{Remaining\%} \\
         \midrule
         \multirow{8}{*}{English} & LAION-en     & 2B       & 280M & 14\% \\
         & LAION-COCO   & 600M     & 300M & 50\% \\
         & DataComp     & 1.4B     & 300M & 21\% \\
         & Coyo         & 700M     & 200M & 28\% \\
         & CC12M        & 12M      & 8M   & 66\% \\
         & CC3M         & 3M       & 3M   & 100\% \\
         & SBU          & 1M       & 0.8M & 80\% \\
         & COCO Caption & 0.6M     & 0.6M & 100\% \\
         \midrule
         \multirow{2}{*}{Chinese} & LAION-zh     & 108M     & 105M & 97\% \\
         & \color{dt}In-house Data & \color{dt}220M & \color{dt}220M & \color{dt}100\% \\
         \midrule
          & Total        & 5B     & 1.4B & 28\% \\
         \bottomrule
    \end{tabular}
    \label{tab:pretraining_data}
\end{table}

We freeze the large language model and only optimize the vision encoder and VL adapter in this stage. The input images are resized to $224 \times 224$. The training objective is to minimize the cross-entropy of the text tokens. The maximum learning rate is $2e^{-4}$ and the training process uses a batch size of 30720 for the image-text pairs, and the entire first stage of pre-training lasts for 50,000 steps, consuming approximately 1.5 billion image-text samples. More hyperparameters are detailed in Appendix \ref{app:hyperparam} and the convergence curve of this stage is shown in Figure \ref{fig:stage1}.

\subsection{Multi-task Pre-training}

In the second stage of multi-task pre-training, we introduce high-quality and fine-grained VL annotation data with a larger input resolution and interleaved image-text data. As summarized in Table~\ref{tab:multitask_data}, we trained Qwen-VL on 7 tasks simultaneously. For text generation, we use the in-house collected corpus to maintain the LLM's ability. Captioning data is the same with Table~\ref{tab:pretraining_data} except for far fewer samples and excluding LAION-COCO. We use a mixture of publicly available data for the VQA task which includes GQA~\citep{gqa}, VGQA~\citep{vg}, VQAv2~\citep{VQAv2}, DVQA~\citep{dvqa}, OCR-VQA~\citep{ocrvqa} and DocVQA~\citep{docvqa}. We follow Kosmos-2 to use the GRIT~\citep{kosmos2} dataset for the grounding task with minor modifications. For the reference grounding and grounded captioning duality tasks, we construct training samples from GRIT~\citep{kosmos2}, Visual Genome~\citep{vg}, RefCOCO~\citep{refcoco}, RefCOCO+, and RefCOCOg~\citep{refcocog}. In order to improve the text-oriented tasks, we collect pdf and HTML format data from Common Crawl\footnote{\scriptsize\url{https://digitalcorpora.org/corpora/file-corpora/cc-main-2021-31-pdf-untruncated}} and generate synthetic OCR data in English and Chinese language with natural scenery background, following~\citep{synthdog}. Finally, we simply construct interleaved image-text data by packing the same task data into sequences of length 2048.

\begin{table}[ht]
    \centering
    \caption{Details of Qwen-VL multi-task pre-training data. 
    }
    \tablestyle{6pt}{1.1}
    \begin{tabular}{l c l}
         \toprule
         \textbf{Task} & \textbf{\# Samples} & \textbf{Dataset} \\
         \midrule
         Captioning     & 19.7M  & \makecell[l]{LAION-en \& zh, DataComp, Coyo, CC12M \& 3M, SBU, \\ COCO, \color{dt}In-house Data} \\
         VQA            & 3.6M  & \makecell[l]{GQA, VGQA, VQAv2, DVQA, OCR-VQA, DocVQA, \\ TextVQA, ChartQA, AI2D} \\
         Grounding\tablefootnote{This task is to generate noun/phrase grounded captions~\citep{kosmos2}.} & 3.5M  & GRIT \\
         Ref Grounding  & 8.7M  & GRIT, Visual Genome, RefCOCO, RefCOCO+, RefCOCOg \\
         Grounded Cap. & 8.7M  & GRIT, Visual Genome, RefCOCO, RefCOCO+, RefCOCOg \\
         OCR            & 24.8M & SynthDoG-en \& zh, Common Crawl pdf \& HTML \\
         Pure-text Autoregression & 7.8M & \color{dt}In-house Data \\
         \bottomrule
    \end{tabular}
    \label{tab:multitask_data}
\end{table}

We increase the input resolution of the visual encoder from $224\times 224$ to $448\times 448$, reducing the information loss caused by image down-sampling. Besides, we ablate the window attention and global attention for higher resolutions of the vision transformer in Appendix \ref{app:window_attention}. We unlocked the large language model and trained the whole model. The training objective is the same as the pre-training stage.

\subsection{Supervised Fine-tuning}
During this stage, we finetuned the Qwen-VL pre-trained model through instruction fine-tuning to enhance its instruction following and dialogue capabilities, resulting in the interactive Qwen-VL-Chat model. The multi-modal instruction tuning data primarily comes from caption data or dialogue data generated through LLM self-instruction, which often only addresses single-image dialogue and reasoning and is limited to image content comprehension. We construct an additional set of dialogue data through manual annotation, model generation, and strategy concatenation to incorporate localization and multi-image comprehension abilities into the Qwen-VL model. We confirm that the model effectively transfers these capabilities to a wider range of languages and question types. Additionally, we mix multi-modal and pure text dialogue data during training to ensure the model's universality in dialogue capabilities. The instruction tuning data amounts to 350k.
In this stage, we freeze the visual encoder and optimize the language model and adapter module. We demonstrate the data format of this stage in Appendix \ref{app:data_format_stage3}.

\section{Evaluation}

In this section, we conduct an overall evaluation on various multi-modal tasks to comprehensively assess our models' visual understanding ability.
In the following, Qwen-VL denotes the model after the multi-task training, and Qwen-VL-Chat denotes the model after supervised fine-tuning (SFT) stage.

Table~\ref{tab:benchmark} provides a detailed summary of the used evaluation benchmarks and corresponding metrics.

\subsection{Image Caption and General Visual Question Answering}

Image caption and general visual question answering (VQA) are two conventional tasks for vision-language models.
Specifically, image caption requires the model to generate a description for a given image and general VQA requires the model to generate an answer for a given image-question pair.

\begin{table}[]
\centering
\caption{Results on Image Captioning and General VQA.}
\scripttablestyle{5pt}{1.05}
\begin{tabular}{@{}l|l|cc|ccccc@{}}
\toprule
\multirow{2}{*}{Model Type} & \multirow{2}{*}{Model} & \multicolumn{2}{c|}{Image Caption} & \multicolumn{5}{c}{General VQA} \\
 &  & \begin{tabular}[c]{@{}c@{}}Nocaps\\ (0-shot)\end{tabular} & \begin{tabular}[c]{@{}c@{}}Flickr30K\\ (0-shot)\end{tabular} & VQAv2 & OKVQA & GQA & \begin{tabular}[c]{@{}c@{}}SciQA-Img\\ (0-shot)\end{tabular} & \begin{tabular}[c]{@{}c@{}}VizWiz\\ (0-shot)\end{tabular} \\ \midrule
\multirow{10}{*}{\begin{tabular}[c]{@{}l@{}}Generalist \\ Models\end{tabular}} & Flamingo-9B & - & 61.5 & 51.8 & 44.7 & - & - & 28.8 \\
 & Flamingo-80B & - & 67.2 & 56.3 & 50.6 & - & - & 31.6 \\
 & Unified-IO-XL & 100.0 & - & 77.9 & 54.0 & - & - & - \\
 & Kosmos-1 & - & 67.1 & 51.0 & - & - & - & 29.2 \\
 & Kosmos-2 & - & 80.5 & 51.1 & - & - & - & - \\
 & BLIP-2 (Vicuna-13B) & 103.9 & 71.6 & 65.0 & 45.9 & 32.3 & 61.0 & 19.6 \\
 & InstructBLIP (Vicuna-13B) & \textbf{121.9} & 82.8 & - & - & 49.5 & 63.1 & 33.4 \\
 & Shikra (Vicuna-13B) & - & 73.9 & 77.36 & 47.16 & - & - & - \\
 & \textbf{Qwen-VL (Qwen-7B)} & 121.4 & \textbf{85.8} & \textbf{79.5} & \textbf{58.6} & \textbf{59.3} & 67.1 & 35.2 \\
 & \textbf{Qwen-VL-Chat} & 120.2 & 81.0 & 78.2 & 56.6 & 57.5 & \textbf{68.2} & \textbf{38.9} \\ \midrule
\begin{tabular}[c]{@{}l@{}}\color{dt}Specialist\\ \color{dt}SOTAs\end{tabular} & \multicolumn{1}{c|}{-} & \begin{tabular}[c]{@{}c@{}}\color{dt}127.0\\ \color{dt}(PALI-17B)\end{tabular} & \begin{tabular}[c]{@{}c@{}}\color{dt}84.5\\ \color{dt}(InstructBLIP\\ \color{dt}-FlanT5-XL)\end{tabular} & \begin{tabular}[c]{@{}c@{}}\color{dt}86.1\\ \color{dt}(PALI-X\\ \color{dt}-55B)\end{tabular} & \begin{tabular}[c]{@{}c@{}}\color{dt}66.1\\ \color{dt}(PALI-X\\ \color{dt}-55B)\end{tabular} & \begin{tabular}[c]{@{}c@{}}\color{dt}72.1\\ \color{dt}(CFR)\end{tabular} & \begin{tabular}[c]{@{}c@{}}\color{dt}92.53\\ \color{dt}(LLaVa+\\ \color{dt}GPT-4)\end{tabular} & \begin{tabular}[c]{@{}c@{}}\color{dt}70.9\\ \color{dt}(PALI-X\\ \color{dt}-55B)\end{tabular} \\ \bottomrule
\end{tabular}
\label{tab:caption_vqa}
\end{table}

For the image caption task, we choose Nocaps \citep{agrawal2019nocaps} and Flickr30K \citep{young2014image_flickr30k} as benchmarks and report CIDEr score~\citep{vedantam2015cider} as metric.
We utilize greedy search for caption generation with a prompt of \textit{"Descripe the image in English:"}.

For general VQA, we utilize five benchmarks including VQAv2~\citep{VQAv2}, OKVQA~\citep{marino2019ok_okvqa}, GQA~\citep{gqa}, ScienceQA (Image Set)~\citep{lu2022learn_scienceqa} and VizWiz VQA~\citep{gurari2018vizwiz}.
For VQAv2, OKVQA, GQA and VizWiz VQA, we employ open-ended answer generation with greedy decoding strategy and a prompt of \textit{"\{question\} Answer:"}, without any constrain on model's output space.
However, for ScienceQA, we constrain the model's output to possible options (instead of open-ended), choose the option with highest confidence as model's prediction, and report the Top-$1$ accuracy.

The overall performance on image caption and general VQA tasks are reported in Table \ref{tab:caption_vqa}.  As the results shown, our Qwen-VL and Qwen-VL-Chat both achieve obviously better results compared to previous generalist models in terms of both two tasks.
Specifically, on zero-shot image caption task, Qwen-VL achieves state-of-the-art performance (\emph{i.e}., 85.8 CIDEr score) on the Flickr30K karpathy-test split, even outperforms previous generalist models with much more parameters (\emph{e.g}., Flamingo-80B with 80B parameters).

On general VQA benchmarks, our models also exhibit distinct advantages compared to others. On VQAv2, OKVQA and GQA benchmarks, Qwen-VL achieves 79.5, 58.6 and 59.3 accuracy respectively, which surpasses recent proposed LVLMs by a large margin.
It's worth noting that Qwen-VL also shows strong zero-shot performance on ScienceQA and VizWiz datasets.

\subsection{Text-oriented Visual Question Answering}

Text-oriented visual understanding has a broad application prospect in real-world scenarios. We assess our models' ability toward text-oriented visual question answering on several benchmarks including TextVQA~\citep{sidorov2020textcaps}, DocVQA~\citep{docvqa}, ChartQA~\citep{masry2022chartqa}, AI2Diagram~\citep{kembhavi2016diagram}, and OCR-VQA~\citep{ocrvqa}.
Similarly, the results are shown in Table~\ref{tab:text_vqa}. Compared to previous generalist models and recent LVLMs, our models show better performance on most benchmarks, frequently by a large margin.

\begin{table}[]
\centering
\tablestyle{4pt}{1.05}
\caption{Results on Text-oriented VQA.}
\begin{tabular}{@{}l|l|ccccc@{}}
\toprule
Model type & Model & TextVQA & DocVQA & ChartQA & AI2D & OCR-VQA \\ \midrule
\multirow{5}{*}{Generalist Models} & BLIP-2 (Vicuna-13B) & 42.4 & - & - & - & - \\
 & InstructBLIP (Vicuna-13B) & 50.7 & - & - & - & - \\
 & mPLUG-DocOwl (LLaMA-7B) & 52.6 & 62.2 & 57.4 & - & - \\
 & Pix2Struct-Large (1.3B) & - & \textbf{76.6} & 58.6 & 42.1 & 71.3 \\
 & \textbf{Qwen-VL (Qwen-7B)} & \textbf{63.8} & 65.1 & 65.7 & \textbf{62.3} & \textbf{75.7} \\
 & \textbf{Qwen-VL-Chat} & 61.5 & 62.6 & \textbf{66.3} & 57.7 & 70.5 \\
 \midrule
\color{dt}Specialist SOTAs & \begin{tabular}[c]{@{}l@{}}\color{dt}PALI-X-55B (Single-task fine-\\\color{dt}tuning, without OCR Pipeline)\end{tabular} & \color{dt}71.44 & \color{dt}80.0 & \color{dt}70.0 & \color{dt}81.2 & \color{dt}75.0 \\ \bottomrule
\end{tabular}
\label{tab:text_vqa}
\end{table}

\subsection{Refer Expression Comprehension}

We show our models' fine-grained image understanding and localization ability by evaluating on a sort of refer expression comprehension benchmarks such as RefCOCO~\citep{refcoco}, RefCOCOg~\citep{refcocog}, RefCOCO+~\citep{refcocog} and GRIT~\citep{gupta2022grit}.
Specifically, the refer expression comprehension task requires the model to localize the target object under the guidance of a description.
The results are shown in Table~\ref{tab:grounding}.
Compared to previous generalist models or recent LVLMs, our models obtain top-tier results on all benchmarks.

\begin{table}[]
\centering
\caption{Results on Referring Expression Comprehension task.}
\tablestyle{3pt}{1.05}
\begin{tabular}{@{}l|l|ccccccccc@{}}
\toprule
\multirow{2}{*}{Model type} & \multirow{2}{*}{Model} & \multicolumn{3}{c}{RefCOCO} & \multicolumn{3}{c}{RefCOCO+} & \multicolumn{2}{c}{RefCOCOg} & GRIT \\
 &  & val & test-A & test-B & val & test-A & test-B & val & test & refexp \\ \midrule
\multirow{8}{*}{Generalist Models} & GPV-2 & - & - & - & - & - & - & - & - & 51.50 \\
 & OFA-L* & 79.96 & 83.67 & 76.39 & 68.29 & 76.00 & 61.75 & 67.57 & 67.58 & 61.70 \\
 & Unified-IO & - & - & - & - & - & - & - & - & \textbf{78.61} \\
 & VisionLLM-H &  & 86.70 & - & - & - & - & - & - & - \\
 & Shikra-7B & 87.01 & 90.61 & 80.24 & 81.60 & 87.36 & 72.12 & 82.27 & 82.19 & 69.34 \\
 & Shikra-13B & 87.83 & 91.11 & 81.81 & 82.89 & 87.79 & 74.41 & 82.64 & 83.16 & 69.03 \\
 & \textbf{Qwen-VL-7B} & \textbf{89.36} & 92.26 & \textbf{85.34} & \textbf{83.12} & 88.25 & \textbf{77.21} & 85.58 & 85.48 & 78.22 \\
 & \textbf{Qwen-VL-7B-Chat} & 88.55 & \textbf{92.27} & 84.51 & 82.82 & \textbf{88.59} & 76.79 & \textbf{85.96} & \textbf{86.32} & - \\ \midrule
\multirow{3}{*}{\color{dt}Specialist SOTAs} & \color{dt}G-DINO-L & \color{dt}90.56 & \color{dt}93.19 & \color{dt}88.24 & \color{dt}82.75 & \color{dt}88.95 & \color{dt}75.92 & \color{dt}86.13 & \color{dt}87.02 & \color{dt}- \\
 & \color{dt}UNINEXT-H & \color{dt}92.64 & \color{dt}94.33 & \color{dt}91.46 & \color{dt}85.24 & \color{dt}89.63 & \color{dt}79.79 & \color{dt}88.73 & \color{dt}89.37 & \color{dt}- \\
 & \color{dt}ONE-PEACE & \color{dt}92.58 & \color{dt}94.18 & \color{dt}89.26 & \color{dt}88.77 & \color{dt}92.21 & \color{dt}83.23 & \color{dt}89.22 & \color{dt}89.27 & \color{dt}- \\
 \bottomrule
\end{tabular}
\label{tab:grounding}
\end{table}

\subsection{Few-shot Learning on Vision-Language Tasks}
\begin{figure*}[ht]
\centering
    \includegraphics[width= 1\textwidth]{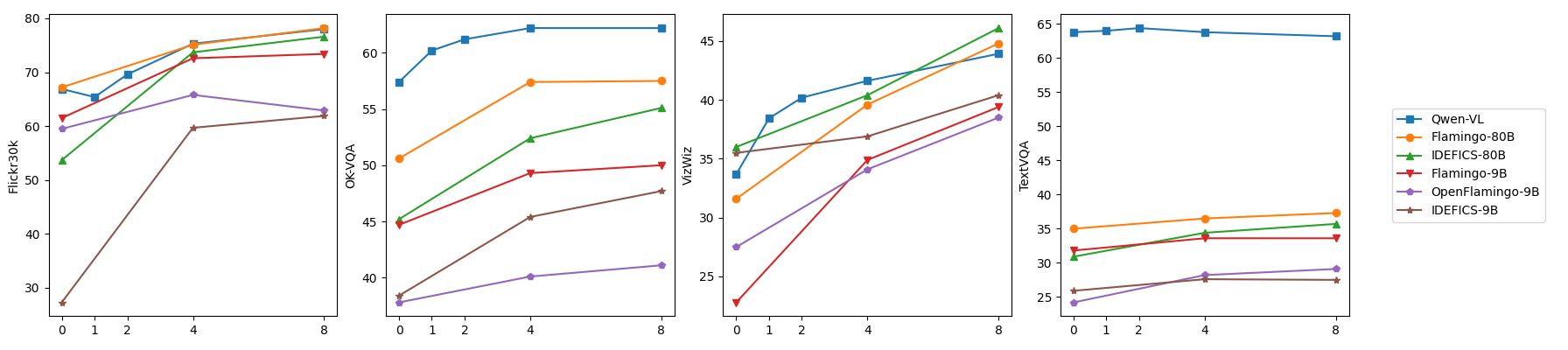}
   \caption{Few-shot learning results of Qwen-VL in comparison with other models.}
\label{fig:fewshot}
\end{figure*}

Our model also exhibits satisfactory in-context learning (\emph{a.k.a.}, few-shot learning) ability. 
As shown in Figure~\ref{fig:fewshot}, Qwen-VL achieves better performance through in-context few-shot learning on OKVQA~\citep{marino2019ok_okvqa}, Vizwiz~\citep{gurari2018vizwiz}, TextVQA~\citep{sidorov2020textcaps}, and Flickr30k~\citep{young2014image_flickr30k} when compared with models with similar number of parameters (Flamingo-9B\citep{alayrac2022flamingo}, OpenFlamingo-9B\citep{awadalla2023openflamingo} and IDEFICS-9B\cite{laurençon2023obelics}). Qwen-VL's performance is even comparable with much larger models (Flamingo-80B and IDEFICS-80B).
Note that we adopt na\"ive random sample to construct the few-shot exemplars, sophisticated few-shot exemplar construction methods such as RICES~\citep{rices} are not used despite better results would be achieved.

\subsection{Instruction Following in Real-world User Behavior}

In addition to previous conventional vision-language evaluations, to evaluate our Qwen-VL-Chat model's capacity under real-world user behavior, we further conduct the evaluations on the TouchStone~\citep{touchstone}, SEED-Bench~\citep{li2023seedbench}, and MME~\citep{fu2023mme}.
TouchStone is an open-ended vision-language instruction-following benchmark.
We compare the instruction-following ability of Qwen-VL-Chat with other instruction-tuned LVLMs in both English and Chinese on the TouchStone benchmark. SEED-Bench consists of 19K multiple-choice questions with accurate human annotations for evaluating Multimodal LLMs, covering 12 evaluation dimensions including both the spatial and temporal understanding. MME measures both perception and cognition abilities on a total of 14 subtasks.

The results on three benchmarks are shown in Table~\ref{tab:intruction_following}.  Qwen-VL-Chat has achieved obvious advantages over other LVLMs on all three datasets, indicating that our model performs better in understanding and answering diverse user instructions. In SEED-Bench, we have found that our model's visual capabilities can be effectively transferred to video tasks by simply sampling four frames.
 In terms of the overall scores presented in TouchStone, our model demonstrates a clear advantage compared to other LVLMs, especially in terms of its Chinese capabilities. In terms of the broad categories of abilities, our model exhibits a more pronounced advantage in understanding and recognition, particularly in areas such as text recognition and chart analysis. For more detailed information, please refer to the TouchStone dataset.

\begin{table}[]
\centering
\tablestyle{5pt}{1.05}
\caption{Results on Instruction-following benchmarks.}
\begin{tabular}{@{}l|ccccccc@{}}
\toprule
\multirow{2}{*}{Model} & \multicolumn{2}{c}{TouchStone} & \multicolumn{3}{c}{SEED-Bench} & \multicolumn{2}{c}{MME}   \\
 &  En&Cn & All & Img &Video & Perception & Cognition \\ \midrule
 VisualGLM & - & 247.1 & - & - & - & 705.31  & 181.79 \\
 PandaGPT & 488.5 & - & - & -&-&642.59 &228.57  \\
 MiniGPT4 & 531.7 &- & 42.8 & 47.4 & 29.9 & 581.67&144.29  \\
 InstructBLIP&552.4 & - & 53.4 & 58.8 & 38.1 & 1212.82&291.79\\
 LLaMA-AdapterV2 &590.1 & - &32.7 & 35.2& 25.8 & 972.67 &248.93\\
 LLaVA  & 602.7 &- & 33.5 & 37.0 & 23.8 & 502.82&214.64 \\
 mPLUG-Owl & 605.4 & - & 34.0 & 37.9 & 23.0 & 967.34 & 276.07 \\ \midrule
 \textbf{Qwen-VL} & - & - & 56.3 & 62.3 &  \textbf{39.1} &  - & -   \\
 \textbf{Qwen-VL-Chat} & \textbf{645.2} & \textbf{401.2} & \textbf{58.2} & \textbf{65.4} & 37.8 &  \textbf{1487.58} & \textbf{360.71}  \\
 \bottomrule
\end{tabular}
\vspace{-0.2cm}
\label{tab:intruction_following}
\end{table}

\section{Related Work}
In recent years, researchers have shown considerable interest in vision-language learning~\citep{vlbert,uniter,oscar,vinvl,unimo,m6,vilt,fiber,xvlm,albef,blip}, especially in the development of multi-task generalist models~\citep{unit,flava,uni-perceiver,coca,wang2022ofa,unified_io,bai2022ofasys}.
CoCa~\citep{coca} proposes an encoder-decoder structure to address image-text retrieval and vision-language generation tasks simultaneously. 
OFA~\citep{wang2022ofa} transforms specific vision-language tasks into sequence-to-sequence tasks using customized task instructions. 
Unified I/O~\citep{unified_io} further introduces more tasks like segmentation and depth estimation into a unified framework.
Another category of research focuses on building vision-language representation models~\citep{clip,align,lit,florence,chinese_clip}.
CLIP~\citep{clip} leverages contrastive learning and large amounts of data to align images and language in a semantic space, resulting in strong generalization capabilities across a wide range of downstream tasks. 
BEIT-3~\citep{beit3} employs a mixture-of-experts (MOE) structure and unified masked token prediction objective, achieving state-of-the-art results on various visual-language tasks. 
In addition to vision-language learning, ImageBind~\citep{imagebind} and ONE-PEACE~\citep{one-peace} align more modalities such as speech into a unified semantic space, thus creating more general representation models.

Despite achieving significant progress, previous vision-language models still have several limitations such as poor robustness in instruction following, limited generalization capabilities in unseen tasks, and a lack of in-context abilities. With the rapid development of large language models (LLMs)~\citep{gpt3,gpt4,anil2023palm,gao2023llama,qwen7b}, researchers have started building more powerful large vision-language models (LVLMs) based on LLMs~\citep{alayrac2022flamingo,chen2022pali, blip2,dai2023instructblip,kosmos,kosmos2,zhu2023minigpt,liu2023visual,ye2023mplug,mPLUG-DocOwl,shikra,li2023otter,videollama,emu}. 
BLIP-2~\citep{blip2} proposes Q-Former to align the frozen vision foundation models and LLMs.
Meanwhile, LLAVA~\citep{liu2023visual} and Mini-GPT4~\citep{zhu2023minigpt} introduce visual instruction tuning to enhance instruction following capabilities in LVLMs. 
Additionally, mPLUG-DocOwl~\citep{mPLUG-DocOwl} incorporates document understanding capabilities into LVLMs by introducing digital documents data. 
Kosmos2~\citep{kosmos2}, Shikra~\citep{shikra}, and BuboGPT~\citep{bubogpt} further enhance LVLMs with visual grounding abilities, enabling region description and localization.
In this work, we integrate image captioning, visual question answering, OCR, document understanding, and visual grounding capabilities into Qwen-VL. The resulting model achieves outstanding performance on these diverse style tasks.

\vspace{-0.2cm}
\section{Conclusion and Future Work}
We release the Qwen-VL series, a set of large-scale multilingual vision-language models that aims to facilitate multimodal research. Qwen-VL outperforms similar models across various benchmarks, supporting multilingual conversations, multi-image interleaved conversations, grounding in Chinese, and fine-grained recognition. 
Moving forward, we are dedicated to further enhancing Qwen-VL's capabilities in several key dimensions: 

\begin{itemize}
\item Integrating Qwen-VL with more modalities, such as speech and video. 
\item Augmenting Qwen-VL by scaling up the model size, training data and higher resolution, enabling it to handle more complex and intricate relationships within multimodal data.
\item Expanding Qwen-VL's prowess in multi-modal generation, specifically in generating high-fidelity images and fluent speech.
\end{itemize}

\bibliographystyle{plainnat} 
\bibliography{references}

\begin{thebibliography}{84}
\providecommand{\natexlab}[1]{#1}
\providecommand{\url}[1]{\texttt{#1}}
\expandafter\ifx\csname urlstyle\endcsname\relax
  \providecommand{\doi}[1]{doi: #1}\else
  \providecommand{\doi}{doi: \begingroup \urlstyle{rm}\Url}\fi

\bibitem[Agrawal et~al.(2019)Agrawal, Desai, Wang, Chen, Jain, Johnson, Batra,
  Parikh, Lee, and Anderson]{agrawal2019nocaps}
Harsh Agrawal, Karan Desai, Yufei Wang, Xinlei Chen, Rishabh Jain, Mark
  Johnson, Dhruv Batra, Devi Parikh, Stefan Lee, and Peter Anderson.
\newblock nocaps: novel object captioning at scale.
\newblock In \emph{ICCV}, 2019.

\bibitem[Alayrac et~al.(2022)Alayrac, Donahue, Luc, Miech, Barr, Hasson, Lenc,
  Mensch, Millican, Reynolds, et~al.]{alayrac2022flamingo}
Jean-Baptiste Alayrac, Jeff Donahue, Pauline Luc, Antoine Miech, Iain Barr,
  Yana Hasson, Karel Lenc, Arthur Mensch, Katherine Millican, Malcolm Reynolds,
  et~al.
\newblock Flamingo: a visual language model for few-shot learning.
\newblock In \emph{NeurIPS}, 2022.

\bibitem[Anil et~al.(2023)Anil, Dai, Firat, Johnson, Lepikhin, Passos, Shakeri,
  Taropa, Bailey, Chen, et~al.]{anil2023palm}
Rohan Anil, Andrew~M Dai, Orhan Firat, Melvin Johnson, Dmitry Lepikhin,
  Alexandre Passos, Siamak Shakeri, Emanuel Taropa, Paige Bailey, Zhifeng Chen,
  et~al.
\newblock Palm 2 technical report.
\newblock \emph{arXiv:2305.10403}, 2023.

\bibitem[Bai et~al.(2022)Bai, Men, Yang, Ren, Dang, Zhang, Zhou, Wang, Tan,
  Yang, et~al.]{bai2022ofasys}
Jinze Bai, Rui Men, Hao Yang, Xuancheng Ren, Kai Dang, Yichang Zhang, Xiaohuan
  Zhou, Peng Wang, Sinan Tan, An~Yang, et~al.
\newblock Ofasys: A multi-modal multi-task learning system for building
  generalist models.
\newblock \emph{arXiv:2212.04408}, 2022.

\bibitem[Bai et~al.(2023)Bai, Yang, Bai, Wang, Zhang, Lin, Wang, Zhou, and
  Zhou]{touchstone}
Shuai Bai, Shusheng Yang, Jinze Bai, Peng Wang, Xingxuan Zhang, Junyang Lin,
  Xinggang Wang, Chang Zhou, and Jingren Zhou.
\newblock Touchstone: Evaluating vision-language models by language models.
\newblock \emph{arXiv:2308.16890}, 2023.

\bibitem[Brown et~al.(2020)Brown, Mann, Ryder, Subbiah, Kaplan, Dhariwal,
  Neelakantan, Shyam, Sastry, Askell, et~al.]{gpt3}
Tom Brown, Benjamin Mann, Nick Ryder, Melanie Subbiah, Jared~D Kaplan, Prafulla
  Dhariwal, Arvind Neelakantan, Pranav Shyam, Girish Sastry, Amanda Askell,
  et~al.
\newblock Language models are few-shot learners.
\newblock In \emph{NeurIPS}, 2020.

\bibitem[Byeon et~al.(2022)Byeon, Park, Kim, Lee, Baek, and Kim]{coyo}
Minwoo Byeon, Beomhee Park, Haecheon Kim, Sungjun Lee, Woonhyuk Baek, and
  Saehoon Kim.
\newblock Coyo-700m: Image-text pair dataset, 2022.
\newblock URL \url{https://github.com/kakaobrain/coyo-dataset}.

\bibitem[Changpinyo et~al.(2021)Changpinyo, Sharma, Ding, and Soricut]{cc12m}
Soravit Changpinyo, Piyush Sharma, Nan Ding, and Radu Soricut.
\newblock Conceptual 12m: Pushing web-scale image-text pre-training to
  recognize long-tail visual concepts.
\newblock In \emph{CVPR}, 2021.

\bibitem[Chen et~al.(2023{\natexlab{a}})Chen, Zhang, Zeng, Zhang, Zhu, and
  Zhao]{shikra}
Keqin Chen, Zhao Zhang, Weili Zeng, Richong Zhang, Feng Zhu, and Rui Zhao.
\newblock Shikra: Unleashing multimodal llm's referential dialogue magic.
\newblock \emph{arXiv:2306.15195}, 2023{\natexlab{a}}.

\bibitem[Chen et~al.(2022)Chen, Wang, Changpinyo, Piergiovanni, Padlewski,
  Salz, Goodman, Grycner, Mustafa, Beyer, et~al.]{chen2022pali}
Xi~Chen, Xiao Wang, Soravit Changpinyo, AJ~Piergiovanni, Piotr Padlewski,
  Daniel Salz, Sebastian Goodman, Adam Grycner, Basil Mustafa, Lucas Beyer,
  et~al.
\newblock Pali: A jointly-scaled multilingual language-image model.
\newblock \emph{arXiv:2209.06794}, 2022.

\bibitem[Chen et~al.(2023{\natexlab{b}})Chen, Djolonga, Padlewski, Mustafa,
  Changpinyo, Wu, Ruiz, Goodman, Wang, Tay, et~al.]{chen2023pali}
Xi~Chen, Josip Djolonga, Piotr Padlewski, Basil Mustafa, Soravit Changpinyo,
  Jialin Wu, Carlos~Riquelme Ruiz, Sebastian Goodman, Xiao Wang, Yi~Tay, et~al.
\newblock Pali-x: On scaling up a multilingual vision and language model.
\newblock \emph{arXiv preprint arXiv:2305.18565}, 2023{\natexlab{b}}.

\bibitem[Chen et~al.(2015)Chen, Fang, Lin, Vedantam, Gupta, Doll{\'a}r, and
  Zitnick]{cococaption}
Xinlei Chen, Hao Fang, Tsung-Yi Lin, Ramakrishna Vedantam, Saurabh Gupta, Piotr
  Doll{\'a}r, and C~Lawrence Zitnick.
\newblock Microsoft coco captions: Data collection and evaluation server.
\newblock \emph{arXiv:1504.00325}, 2015.

\bibitem[Chen et~al.(2020)Chen, Li, Yu, Kholy, Ahmed, Gan, Cheng, and
  Liu]{uniter}
Yen-Chun Chen, Linjie Li, Licheng Yu, Ahmed~El Kholy, Faisal Ahmed, Zhe Gan,
  Yu~Cheng, and Jingjing Liu.
\newblock Uniter: Universal image-text representation learning.
\newblock In \emph{ECCV}, 2020.

\bibitem[Dai et~al.(2023)Dai, Li, Li, Tiong, Zhao, Wang, Li, Fung, and
  Hoi]{dai2023instructblip}
Wenliang Dai, Junnan Li, Dongxu Li, Anthony Meng~Huat Tiong, Junqi Zhao,
  Weisheng Wang, Boyang Li, Pascale Fung, and Steven Hoi.
\newblock Instructblip: Towards general-purpose vision-language models with
  instruction tuning.
\newblock \emph{arXiv:2305.06500}, 2023.

\bibitem[Dosovitskiy et~al.(2021)Dosovitskiy, Beyer, Kolesnikov, Weissenborn,
  Zhai, Unterthiner, Dehghani, Minderer, Heigold, Gelly, Uszkoreit, and
  Houlsby]{dosovitskiy2020vit}
Alexey Dosovitskiy, Lucas Beyer, Alexander Kolesnikov, Dirk Weissenborn,
  Xiaohua Zhai, Thomas Unterthiner, Mostafa Dehghani, Matthias Minderer, Georg
  Heigold, Sylvain Gelly, Jakob Uszkoreit, and Neil Houlsby.
\newblock An image is worth 16x16 words: Transformers for image recognition at
  scale.
\newblock In \emph{ICLR}, 2021.

\bibitem[Dou et~al.(2022)Dou, Kamath, Gan, Zhang, Wang, Li, Liu, Liu, LeCun,
  Peng, Gao, and Wang]{fiber}
Zi-Yi* Dou, Aishwarya* Kamath, Zhe* Gan, Pengchuan Zhang, Jianfeng Wang, Linjie
  Li, Zicheng Liu, Ce~Liu, Yann LeCun, Nanyun Peng, Jianfeng Gao, and Lijuan
  Wang.
\newblock Coarse-to-fine vision-language pre-training with fusion in the
  backbone.
\newblock In \emph{NeurIPS}, 2022.

\bibitem[Fu et~al.(2023)Fu, Chen, Shen, Qin, Zhang, Lin, Qiu, Lin, Yang, Zheng,
  et~al.]{fu2023mme}
Chaoyou Fu, Peixian Chen, Yunhang Shen, Yulei Qin, Mengdan Zhang, Xu~Lin,
  Zhenyu Qiu, Wei Lin, Jinrui Yang, Xiawu Zheng, et~al.
\newblock Mme: A comprehensive evaluation benchmark for multimodal large
  language models.
\newblock \emph{arXiv:2306.13394}, 2023.

\bibitem[Gadre et~al.(2023)Gadre, Ilharco, Fang, Hayase, Smyrnis, Nguyen,
  Marten, Wortsman, Ghosh, Zhang, et~al.]{datacomp}
Samir~Yitzhak Gadre, Gabriel Ilharco, Alex Fang, Jonathan Hayase, Georgios
  Smyrnis, Thao Nguyen, Ryan Marten, Mitchell Wortsman, Dhruba Ghosh, Jieyu
  Zhang, et~al.
\newblock Datacomp: In search of the next generation of multimodal datasets.
\newblock \emph{arXiv:2304.14108}, 2023.

\bibitem[Gao et~al.(2023)Gao, Han, Zhang, Lin, Geng, Zhou, Zhang, Lu, He, Yue,
  et~al.]{gao2023llama}
Peng Gao, Jiaming Han, Renrui Zhang, Ziyi Lin, Shijie Geng, Aojun Zhou, Wei
  Zhang, Pan Lu, Conghui He, Xiangyu Yue, et~al.
\newblock Llama-adapter v2: Parameter-efficient visual instruction model.
\newblock \emph{arXiv:2304.15010}, 2023.

\bibitem[Girdhar et~al.(2023)Girdhar, El-Nouby, Liu, Singh, Alwala, Joulin, and
  Misra]{imagebind}
Rohit Girdhar, Alaaeldin El-Nouby, Zhuang Liu, Mannat Singh, Kalyan~Vasudev
  Alwala, Armand Joulin, and Ishan Misra.
\newblock Imagebind: One embedding space to bind them all.
\newblock In \emph{CVPR}, 2023.

\bibitem[Google(2023)]{puppeteer}
Google.
\newblock Puppeteer, 2023.
\newblock URL \url{https://github.com/puppeteer/puppeteer}.

\bibitem[Goyal et~al.(2017)Goyal, Khot, Summers-Stay, Batra, and Parikh]{VQAv2}
Yash Goyal, Tejas Khot, Douglas Summers-Stay, Dhruv Batra, and Devi Parikh.
\newblock Making the v in vqa matter: Elevating the role of image understanding
  in visual question answering.
\newblock In \emph{CVPR}, 2017.

\bibitem[Gupta et~al.(2022)Gupta, Marten, Kembhavi, and Hoiem]{gupta2022grit}
Tanmay Gupta, Ryan Marten, Aniruddha Kembhavi, and Derek Hoiem.
\newblock Grit: General robust image task benchmark.
\newblock \emph{arXiv:2204.13653}, 2022.

\bibitem[Gurari et~al.(2018)Gurari, Li, Stangl, Guo, Lin, Grauman, Luo, and
  Bigham]{gurari2018vizwiz}
Danna Gurari, Qing Li, Abigale~J Stangl, Anhong Guo, Chi Lin, Kristen Grauman,
  Jiebo Luo, and Jeffrey~P Bigham.
\newblock Vizwiz grand challenge: Answering visual questions from blind people.
\newblock In \emph{CVPR}, 2018.

\bibitem[Hu and Singh(2021)]{unit}
Ronghang Hu and Amanpreet Singh.
\newblock Unit: Multimodal multitask learning with a unified transformer.
\newblock In \emph{ICCV}, 2021.

\bibitem[Huang et~al.(2023)Huang, Dong, Wang, Hao, Singhal, Ma, Lv, Cui,
  Mohammed, Liu, et~al.]{kosmos}
Shaohan Huang, Li~Dong, Wenhui Wang, Yaru Hao, Saksham Singhal, Shuming Ma,
  Tengchao Lv, Lei Cui, Owais~Khan Mohammed, Qiang Liu, et~al.
\newblock Language is not all you need: Aligning perception with language
  models.
\newblock \emph{arXiv:2302.14045}, 2023.

\bibitem[Hudson and Manning(2019)]{gqa}
Drew~A Hudson and Christopher~D Manning.
\newblock Gqa: A new dataset for real-world visual reasoning and compositional
  question answering.
\newblock In \emph{CVPR}, 2019.

\bibitem[Ilharco et~al.(2021)Ilharco, Wortsman, Wightman, Gordon, Carlini,
  Taori, Dave, Shankar, Namkoong, Miller, Hajishirzi, Farhadi, and
  Schmidt]{openclip}
Gabriel Ilharco, Mitchell Wortsman, Ross Wightman, Cade Gordon, Nicholas
  Carlini, Rohan Taori, Achal Dave, Vaishaal Shankar, Hongseok Namkoong, John
  Miller, Hannaneh Hajishirzi, Ali Farhadi, and Ludwig Schmidt.
\newblock Openclip, 2021.
\newblock URL \url{https://doi.org/10.5281/zenodo.5143773}.

\bibitem[Jia et~al.(2021)Jia, Yang, Xia, Chen, Parekh, Pham, Le, Sung, Li, and
  Duerig]{align}
Chao Jia, Yinfei Yang, Ye~Xia, Yi-Ting Chen, Zarana Parekh, Hieu Pham, Quoc~V
  Le, Yunhsuan Sung, Zhen Li, and Tom Duerig.
\newblock Scaling up visual and vision-language representation learning with
  noisy text supervision.
\newblock \emph{arXiv:2102.05918}, 2021.

\bibitem[Kafle et~al.(2018)Kafle, Price, Cohen, and Kanan]{dvqa}
Kushal Kafle, Brian Price, Scott Cohen, and Christopher Kanan.
\newblock Dvqa: Understanding data visualizations via question answering.
\newblock In \emph{CVPR}, 2018.

\bibitem[Kazemzadeh et~al.(2014)Kazemzadeh, Ordonez, Matten, and Berg]{refcoco}
Sahar Kazemzadeh, Vicente Ordonez, Mark Matten, and Tamara Berg.
\newblock Referitgame: Referring to objects in photographs of natural scenes.
\newblock In \emph{EMNLP}, 2014.

\bibitem[Kembhavi et~al.(2016)Kembhavi, Salvato, Kolve, Seo, Hajishirzi, and
  Farhadi]{kembhavi2016diagram}
Aniruddha Kembhavi, Mike Salvato, Eric Kolve, Minjoon Seo, Hannaneh Hajishirzi,
  and Ali Farhadi.
\newblock A diagram is worth a dozen images.
\newblock In \emph{ECCV}, 2016.

\bibitem[Kim et~al.(2022)Kim, Hong, Yim, Nam, Park, Yim, Hwang, Yun, Han, and
  Park]{synthdog}
Geewook Kim, Teakgyu Hong, Moonbin Yim, JeongYeon Nam, Jinyoung Park, Jinyeong
  Yim, Wonseok Hwang, Sangdoo Yun, Dongyoon Han, and Seunghyun Park.
\newblock Ocr-free document understanding transformer.
\newblock In \emph{ECCV}, 2022.

\bibitem[Kim et~al.(2021)Kim, Son, and Kim]{vilt}
Wonjae Kim, Bokyung Son, and Ildoo Kim.
\newblock Vilt: Vision-and-language transformer without convolution or region
  supervision.
\newblock In \emph{ICML}, 2021.

\bibitem[Krishna et~al.(2017)Krishna, Zhu, Groth, Johnson, Hata, Kravitz, Chen,
  Kalantidis, Li, Shamma, et~al.]{vg}
Ranjay Krishna, Yuke Zhu, Oliver Groth, Justin Johnson, Kenji Hata, Joshua
  Kravitz, Stephanie Chen, Yannis Kalantidis, Li-Jia Li, David~A Shamma, et~al.
\newblock Visual genome: Connecting language and vision using crowdsourced
  dense image annotations.
\newblock In \emph{IJCV}, 2017.

\bibitem[Li et~al.(2023{\natexlab{a}})Li, Zhang, Chen, Wang, Yang, and
  Liu]{li2023otter}
Bo~Li, Yuanhan Zhang, Liangyu Chen, Jinghao Wang, Jingkang Yang, and Ziwei Liu.
\newblock Otter: A multi-modal model with in-context instruction tuning.
\newblock \emph{arXiv:2305.03726}, 2023{\natexlab{a}}.

\bibitem[Li et~al.(2023{\natexlab{b}})Li, Wang, Wang, Ge, Ge, and
  Shan]{li2023seedbench}
Bohao Li, Rui Wang, Guangzhi Wang, Yuying Ge, Yixiao Ge, and Ying Shan.
\newblock Seed-bench: Benchmarking multimodal llms with generative
  comprehension.
\newblock \emph{arXiv:2307.16125}, 2023{\natexlab{b}}.

\bibitem[Li et~al.(2021{\natexlab{a}})Li, Selvaraju, Gotmare, Joty, Xiong, and
  Hoi]{albef}
Junnan Li, Ramprasaath~R Selvaraju, Akhilesh~Deepak Gotmare, Shafiq Joty,
  Caiming Xiong, and Steven Hoi.
\newblock Align before fuse: Vision and language representation learning with
  momentum distillation.
\newblock In \emph{NeurIPS}, 2021{\natexlab{a}}.

\bibitem[Li et~al.(2022)Li, Li, Xiong, and Hoi]{blip}
Junnan Li, Dongxu Li, Caiming Xiong, and Steven C.~H. Hoi.
\newblock Blip: Bootstrapping language-image pre-training for unified
  vision-language understanding and generation.
\newblock In \emph{ICML}, 2022.

\bibitem[Li et~al.(2023{\natexlab{c}})Li, Li, Savarese, and Hoi]{blip2}
Junnan Li, Dongxu Li, Silvio Savarese, and Steven Hoi.
\newblock Blip-2: Bootstrapping language-image pre-training with frozen image
  encoders and large language models.
\newblock \emph{arXiv:2301.12597}, 2023{\natexlab{c}}.

\bibitem[Li et~al.(2021{\natexlab{b}})Li, Gao, Niu, Xiao, Liu, Liu, Wu, and
  Wang]{unimo}
Wei Li, Can Gao, Guocheng Niu, Xinyan Xiao, Hao Liu, Jiachen Liu, Hua Wu, and
  Haifeng Wang.
\newblock {UNIMO:} towards unified-modal understanding and generation via
  cross-modal contrastive learning.
\newblock In \emph{ACL}, 2021{\natexlab{b}}.

\bibitem[Li et~al.(2020)Li, Yin, Li, Hu, Zhang, Zhang, Wang, Hu, Dong, Wei,
  Choi, and Gao]{oscar}
Xiujun Li, Xi~Yin, Chunyuan Li, Xiaowei Hu, Pengchuan Zhang, Lei Zhang, Lijuan
  Wang, Houdong Hu, Li~Dong, Furu Wei, Yejin Choi, and Jianfeng Gao.
\newblock Oscar: Object-semantics aligned pre-training for vision-language
  tasks.
\newblock In \emph{ECCV}, 2020.

\bibitem[Lin et~al.(2021)Lin, Men, Yang, Zhou, Ding, Zhang, Wang, Wang, Jiang,
  Jia, et~al.]{m6}
Junyang Lin, Rui Men, An~Yang, Chang Zhou, Ming Ding, Yichang Zhang, Peng Wang,
  Ang Wang, Le~Jiang, Xianyan Jia, et~al.
\newblock M6: A chinese multimodal pretrainer.
\newblock In \emph{KDD}, 2021.

\bibitem[Lin et~al.(2014)Lin, Maire, Belongie, Hays, Perona, Ramanan,
  Doll{\'a}r, and Zitnick]{lin2014microsoft}
Tsung-Yi Lin, Michael Maire, Serge Belongie, James Hays, Pietro Perona, Deva
  Ramanan, Piotr Doll{\'a}r, and C~Lawrence Zitnick.
\newblock Microsoft coco: Common objects in context.
\newblock In \emph{ECCV}, 2014.

\bibitem[Liu et~al.(2023)Liu, Li, Wu, and Lee]{liu2023visual}
Haotian Liu, Chunyuan Li, Qingyang Wu, and Yong~Jae Lee.
\newblock Visual instruction tuning.
\newblock \emph{arXiv:2304.08485}, 2023.

\bibitem[Lu et~al.(2022{\natexlab{a}})Lu, Clark, Zellers, Mottaghi, and
  Kembhavi]{unified_io}
Jiasen Lu, Christopher Clark, Rowan Zellers, Roozbeh Mottaghi, and Aniruddha
  Kembhavi.
\newblock Unified-io: A unified model for vision, language, and multi-modal
  tasks.
\newblock \emph{arXiv:2206.08916}, 2022{\natexlab{a}}.

\bibitem[Lu et~al.(2022{\natexlab{b}})Lu, Mishra, Xia, Qiu, Chang, Zhu,
  Tafjord, Clark, and Kalyan]{lu2022learn_scienceqa}
Pan Lu, Swaroop Mishra, Tanglin Xia, Liang Qiu, Kai-Wei Chang, Song-Chun Zhu,
  Oyvind Tafjord, Peter Clark, and Ashwin Kalyan.
\newblock Learn to explain: Multimodal reasoning via thought chains for science
  question answering.
\newblock In \emph{NeurIPS}, 2022{\natexlab{b}}.

\bibitem[Mao et~al.(2016)Mao, Huang, Toshev, Camburu, Yuille, and
  Murphy]{refcocog}
Junhua Mao, Jonathan Huang, Alexander Toshev, Oana Camburu, Alan~L Yuille, and
  Kevin Murphy.
\newblock Generation and comprehension of unambiguous object descriptions.
\newblock In \emph{CVPR}, 2016.

\bibitem[Marino et~al.(2019)Marino, Rastegari, Farhadi, and
  Mottaghi]{marino2019ok_okvqa}
Kenneth Marino, Mohammad Rastegari, Ali Farhadi, and Roozbeh Mottaghi.
\newblock Ok-vqa: A visual question answering benchmark requiring external
  knowledge.
\newblock In \emph{CVPR}, 2019.

\bibitem[Masry et~al.(2022)Masry, Long, Tan, Joty, and Hoque]{masry2022chartqa}
Ahmed Masry, Do~Xuan Long, Jia~Qing Tan, Shafiq Joty, and Enamul Hoque.
\newblock Chartqa: A benchmark for question answering about charts with visual
  and logical reasoning.
\newblock \emph{arXiv:2203.10244}, 2022.

\bibitem[Mathew et~al.(2021)Mathew, Karatzas, and Jawahar]{docvqa}
Minesh Mathew, Dimosthenis Karatzas, and CV~Jawahar.
\newblock Docvqa: A dataset for vqa on document images.
\newblock In \emph{WACV}, 2021.

\bibitem[Mishra et~al.(2019)Mishra, Shekhar, Singh, and Chakraborty]{ocrvqa}
Anand Mishra, Shashank Shekhar, Ajeet~Kumar Singh, and Anirban Chakraborty.
\newblock Ocr-vqa: Visual question answering by reading text in images.
\newblock In \emph{ICDAR}, 2019.

\bibitem[Openai()]{chatml}
Openai.
\newblock Chatml documents.
\newblock URL
  \url{https://github.com/openai/openai-python/blob/main/chatml.md}.

\bibitem[OpenAI(2023)]{gpt4}
OpenAI.
\newblock Gpt-4 technical report, 2023.

\bibitem[Ordonez et~al.(2011)Ordonez, Kulkarni, and Berg]{sbu}
Vicente Ordonez, Girish Kulkarni, and Tamara Berg.
\newblock Im2text: Describing images using 1 million captioned photographs.
\newblock In \emph{NeurIPS}, 2011.

\bibitem[Peng et~al.(2023)Peng, Wang, Dong, Hao, Huang, Ma, and Wei]{kosmos2}
Zhiliang Peng, Wenhui Wang, Li~Dong, Yaru Hao, Shaohan Huang, Shuming Ma, and
  Furu Wei.
\newblock Kosmos-2: Grounding multimodal large language models to the world.
\newblock \emph{arXiv:2306.14824}, 2023.

\bibitem[Qwen(2023)]{qwen7b}
Qwen.
\newblock Introducing qwen-7b: Open foundation and human-aligned models (of the
  state-of-the-arts), 2023.
\newblock URL \url{https://github.com/QwenLM/Qwen-7B}.

\bibitem[Radford et~al.(2021)Radford, Kim, Hallacy, Ramesh, Goh, Agarwal,
  Sastry, Askell, Mishkin, Clark, et~al.]{clip}
Alec Radford, Jong~Wook Kim, Chris Hallacy, Aditya Ramesh, Gabriel Goh,
  Sandhini Agarwal, Girish Sastry, Amanda Askell, Pamela Mishkin, Jack Clark,
  et~al.
\newblock Learning transferable visual models from natural language
  supervision.
\newblock In \emph{ICML}, 2021.

\bibitem[Schuhmann et~al.(2022{\natexlab{a}})Schuhmann, Beaumont, Vencu,
  Gordon, Wightman, Cherti, Coombes, Katta, Mullis, Wortsman, et~al.]{laion5b}
Christoph Schuhmann, Romain Beaumont, Richard Vencu, Cade Gordon, Ross
  Wightman, Mehdi Cherti, Theo Coombes, Aarush Katta, Clayton Mullis, Mitchell
  Wortsman, et~al.
\newblock Laion-5b: An open large-scale dataset for training next generation
  image-text models.
\newblock \emph{arXiv:2210.08402}, 2022{\natexlab{a}}.

\bibitem[Schuhmann et~al.(2022{\natexlab{b}})Schuhmann, Köpf, Vencu, Coombes,
  and Beaumont]{laioncoco}
Christoph Schuhmann, Andreas Köpf, Richard Vencu, Theo Coombes, and Romain
  Beaumont.
\newblock Laion coco: 600m synthetic captions from laion2b-en.
\newblock \emph{https://laion.ai/blog/laion-coco/}, 2022{\natexlab{b}}.

\bibitem[Sharma et~al.(2018)Sharma, Ding, Goodman, and Soricut]{cc3m}
Piyush Sharma, Nan Ding, Sebastian Goodman, and Radu Soricut.
\newblock Conceptual captions: A cleaned, hypernymed, image alt-text dataset
  for automatic image captioning.
\newblock In \emph{ACL}, 2018.

\bibitem[Sidorov et~al.(2020)Sidorov, Hu, Rohrbach, and
  Singh]{sidorov2020textcaps}
Oleksii Sidorov, Ronghang Hu, Marcus Rohrbach, and Amanpreet Singh.
\newblock Textcaps: a dataset for image captioning with reading comprehension.
\newblock In \emph{ECCV}, 2020.

\bibitem[Singh et~al.(2022)Singh, Hu, Goswami, Couairon, Galuba, Rohrbach, and
  Kiela]{flava}
Amanpreet Singh, Ronghang Hu, Vedanuj Goswami, Guillaume Couairon, Wojciech
  Galuba, Marcus Rohrbach, and Douwe Kiela.
\newblock Flava: A foundational language and vision alignment model.
\newblock In \emph{CVPR}, 2022.

\bibitem[Software(2015)]{pymupdf}
Artifex Software.
\newblock Pymupdf, 2015.
\newblock URL \url{https://github.com/pymupdf/PyMuPDF}.

\bibitem[Su et~al.(2019)Su, Zhu, Cao, Li, Lu, Wei, and Dai]{vlbert}
Weijie Su, Xizhou Zhu, Yue Cao, Bin Li, Lewei Lu, Furu Wei, and Jifeng Dai.
\newblock Vl-bert: Pre-training of generic visual-linguistic representations.
\newblock In \emph{ICLR}, 2019.

\bibitem[Sun et~al.(2023)Sun, Yu, Cui, Zhang, Zhang, Wang, Gao, Liu, Huang, and
  Wang]{emu}
Quan Sun, Qiying Yu, Yufeng Cui, Fan Zhang, Xiaosong Zhang, Yueze Wang,
  Hongcheng Gao, Jingjing Liu, Tiejun Huang, and Xinlong Wang.
\newblock Generative pretraining in multimodality.
\newblock \emph{arXiv:2307.05222}, 2023.

\bibitem[Vedantam et~al.(2015)Vedantam, Lawrence~Zitnick, and
  Parikh]{vedantam2015cider}
Ramakrishna Vedantam, C~Lawrence~Zitnick, and Devi Parikh.
\newblock Cider: Consensus-based image description evaluation.
\newblock In \emph{CVPR}, 2015.

\bibitem[Wang et~al.(2022{\natexlab{a}})Wang, Yang, Men, Lin, Bai, Li, Ma,
  Zhou, Zhou, and Yang]{wang2022ofa}
Peng Wang, An~Yang, Rui Men, Junyang Lin, Shuai Bai, Zhikang Li, Jianxin Ma,
  Chang Zhou, Jingren Zhou, and Hongxia Yang.
\newblock Ofa: Unifying architectures, tasks, and modalities through a simple
  sequence-to-sequence learning framework.
\newblock In \emph{ICML}, 2022{\natexlab{a}}.

\bibitem[Wang et~al.(2023)Wang, Wang, Lin, Bai, Zhou, Zhou, Wang, and
  Zhou]{one-peace}
Peng Wang, Shijie Wang, Junyang Lin, Shuai Bai, Xiaohuan Zhou, Jingren Zhou,
  Xinggang Wang, and Chang Zhou.
\newblock One-peace: Exploring one general representation model toward
  unlimited modalities.
\newblock \emph{arXiv:2305.11172}, 2023.

\bibitem[Wang et~al.(2022{\natexlab{b}})Wang, Bao, Dong, Bjorck, Peng, Liu,
  Aggarwal, Mohammed, Singhal, Som, et~al.]{beit3}
Wenhui Wang, Hangbo Bao, Li~Dong, Johan Bjorck, Zhiliang Peng, Qiang Liu, Kriti
  Aggarwal, Owais~Khan Mohammed, Saksham Singhal, Subhojit Som, et~al.
\newblock Image as a foreign language: Beit pretraining for all vision and
  vision-language tasks.
\newblock \emph{arXiv:2208.10442}, 2022{\natexlab{b}}.

\bibitem[Yang et~al.(2022{\natexlab{a}})Yang, Pan, Lin, Men, Zhang, Zhou, and
  Zhou]{chinese_clip}
An~Yang, Junshu Pan, Junyang Lin, Rui Men, Yichang Zhang, Jingren Zhou, and
  Chang Zhou.
\newblock Chinese clip: Contrastive vision-language pretraining in chinese.
\newblock \emph{arXiv:2211.01335}, 2022{\natexlab{a}}.

\bibitem[Yang et~al.(2022{\natexlab{b}})Yang, Gan, Wang, Hu, Lu, Liu, and
  Wang]{rices}
Zhengyuan Yang, Zhe Gan, Jianfeng Wang, Xiaowei Hu, Yumao Lu, Zicheng Liu, and
  Lijuan Wang.
\newblock An empirical study of gpt-3 for few-shot knowledge-based vqa.
\newblock In \emph{AAAI}, 2022{\natexlab{b}}.

\bibitem[Ye et~al.(2023{\natexlab{a}})Ye, Hu, Xu, Ye, Yan, Dan, Zhao, Xu, Li,
  Tian, et~al.]{mPLUG-DocOwl}
Jiabo Ye, Anwen Hu, Haiyang Xu, Qinghao Ye, Ming Yan, Yuhao Dan, Chenlin Zhao,
  Guohai Xu, Chenliang Li, Junfeng Tian, et~al.
\newblock mplug-docowl: Modularized multimodal large language model for
  document understanding.
\newblock \emph{arXiv:2307.02499}, 2023{\natexlab{a}}.

\bibitem[Ye et~al.(2023{\natexlab{b}})Ye, Xu, Xu, Ye, Yan, Zhou, Wang, Hu, Shi,
  Shi, et~al.]{ye2023mplug}
Qinghao Ye, Haiyang Xu, Guohai Xu, Jiabo Ye, Ming Yan, Yiyang Zhou, Junyang
  Wang, Anwen Hu, Pengcheng Shi, Yaya Shi, et~al.
\newblock mplug-owl: Modularization empowers large language models with
  multimodality.
\newblock \emph{arXiv:2304.14178}, 2023{\natexlab{b}}.

\bibitem[Young et~al.(2014)Young, Lai, Hodosh, and
  Hockenmaier]{young2014image_flickr30k}
Peter Young, Alice Lai, Micah Hodosh, and Julia Hockenmaier.
\newblock From image descriptions to visual denotations: New similarity metrics
  for semantic inference over event descriptions.
\newblock In \emph{ACL}, 2014.

\bibitem[Yu et~al.(2022)Yu, Wang, Vasudevan, Yeung, Seyedhosseini, and
  Wu]{coca}
Jiahui Yu, Zirui Wang, Vijay Vasudevan, Legg Yeung, Mojtaba Seyedhosseini, and
  Yonghui Wu.
\newblock Coca: Contrastive captioners are image-text foundation models.
\newblock \emph{arXiv:2205.01917}, 2022.

\bibitem[Yuan et~al.(2021)Yuan, Chen, Chen, Codella, Dai, Gao, Hu, Huang, Li,
  Li, Liu, Liu, Liu, Lu, Shi, Wang, Wang, Xiao, Xiao, Yang, Zeng, Zhou, and
  Zhang]{florence}
Lu~Yuan, Dongdong Chen, Yi-Ling Chen, Noel C.~F. Codella, Xiyang Dai, Jianfeng
  Gao, Houdong Hu, Xuedong Huang, Boxin Li, Chunyuan Li, Ce~Liu, Mengchen Liu,
  Zicheng Liu, Yumao Lu, Yu~Shi, Lijuan Wang, Jianfeng Wang, Bin Xiao, Zhen
  Xiao, Jianwei Yang, Michael Zeng, Luowei Zhou, and Pengchuan Zhang.
\newblock Florence: A new foundation model for computer vision.
\newblock \emph{arXiv:2111.11432}, 2021.

\bibitem[Zeng et~al.(2021)Zeng, Zhang, and Li]{xvlm}
Yan Zeng, Xinsong Zhang, and Hang Li.
\newblock Multi-grained vision language pre-training: Aligning texts with
  visual concepts.
\newblock \emph{arXiv:2111.08276}, 2021.

\bibitem[Zhai et~al.(2022)Zhai, Wang, Mustafa, Steiner, Keysers, Kolesnikov,
  and Beyer]{lit}
Xiaohua Zhai, Xiao Wang, Basil Mustafa, Andreas Steiner, Daniel Keysers,
  Alexander Kolesnikov, and Lucas Beyer.
\newblock Lit: Zero-shot transfer with locked-image text tuning.
\newblock In \emph{CVPR}, 2022.

\bibitem[Zhang et~al.(2023)Zhang, Li, and Bing]{videollama}
Hang Zhang, Xin Li, and Lidong Bing.
\newblock Video-llama: An instruction-tuned audio-visual language model for
  video understanding.
\newblock \emph{arXiv:2306.02858}, 2023.

\bibitem[Zhang et~al.(2021)Zhang, Li, Hu, Yang, Zhang, Wang, Choi, and
  Gao]{vinvl}
Pengchuan Zhang, Xiujun Li, Xiaowei Hu, Jianwei Yang, Lei Zhang, Lijuan Wang,
  Yejin Choi, and Jianfeng Gao.
\newblock Vinvl: Revisiting visual representations in vision-language models.
\newblock In \emph{CVPR}, 2021.

\bibitem[Zhao et~al.(2023)Zhao, Lin, Zhou, Huang, Feng, and Kang]{bubogpt}
Yang Zhao, Zhijie Lin, Daquan Zhou, Zilong Huang, Jiashi Feng, and Bingyi Kang.
\newblock Bubogpt: Enabling visual grounding in multi-modal llms.
\newblock \emph{arXiv:2307.08581}, 2023.

\bibitem[Zhu et~al.(2023)Zhu, Chen, Shen, Li, and Elhoseiny]{zhu2023minigpt}
Deyao Zhu, Jun Chen, Xiaoqian Shen, Xiang Li, and Mohamed Elhoseiny.
\newblock Minigpt-4: Enhancing vision-language understanding with advanced
  large language models.
\newblock \emph{arXiv:2304.10592}, 2023.

\bibitem[Zhu et~al.(2022)Zhu, Zhu, Li, Wu, Li, Wang, and Dai]{uni-perceiver}
Xizhou Zhu, Jinguo Zhu, Hao Li, Xiaoshi Wu, Hongsheng Li, Xiaohua Wang, and
  Jifeng Dai.
\newblock Uni-perceiver: Pre-training unified architecture for generic
  perception for zero-shot and few-shot tasks.
\newblock In \emph{CVPR}, 2022.

\end{thebibliography}

%%%%%%%%%%%%%%%%%%%%%%%%%%%%%%%%%%%%%%%%%%%%%%%%%%%%%%%%%%%%

\clearpage
\newpage

\appendix
% \section{Appendix}
\section{Dataset details}
\label{app:dataset}

\subsection{Image-text pairs}
We use web-crawled image-text pairs dataset for pre-training, which includes LAION-en~\citep{laion5b}, LAION-zh~\citep{laion5b}, LAION-COCO~\citep{laioncoco}, DataComp~\citep{datacomp} and Coyo~\citep{coyo}. We clean these noisy data by several steps:
\begin{enumerate}
    \item Removing pairs with too large aspect ratio of the image
    \item Removing pairs with too small image
    \item Removing pairs with a harsh CLIP score (dataset-specific)
    \item Removing pairs with text containing non-English or non-Chinese characters
    \item Removing pairs with text containing emoji characters
    \item Removing pairs with text length too short or too long
    \item Cleaning the text's HTML-tagged part
    \item Cleaning the text with certain unregular patterns
\end{enumerate}

For academic caption datasets, we remove pairs whose text contains the special tags in CC12M~\citep{cc12m} and SBU~\citep{sbu}. If there is more than one text matching the same image, we select the longest one.

\subsection{VQA}
For the VQAv2~\citep{VQAv2} dataset, we select the answer annotation based on the maximum confidence. For other VQA datasets, we didn't do anything special.

\subsection{Grounding}
For the GRIT~\citep{kosmos2} dataset, we found that there are many recursive grounding box labels in one caption. We use the greedy algorithm to clean the caption to make sure each image contains the most box labels with no recursive box labels. For other grounding datasets, we simply concatenate the noun/phrase with respective bounding box coordinates.

\subsection{OCR}

\begin{figure*}[htp]
\centering
\includegraphics[width= 1\textwidth]{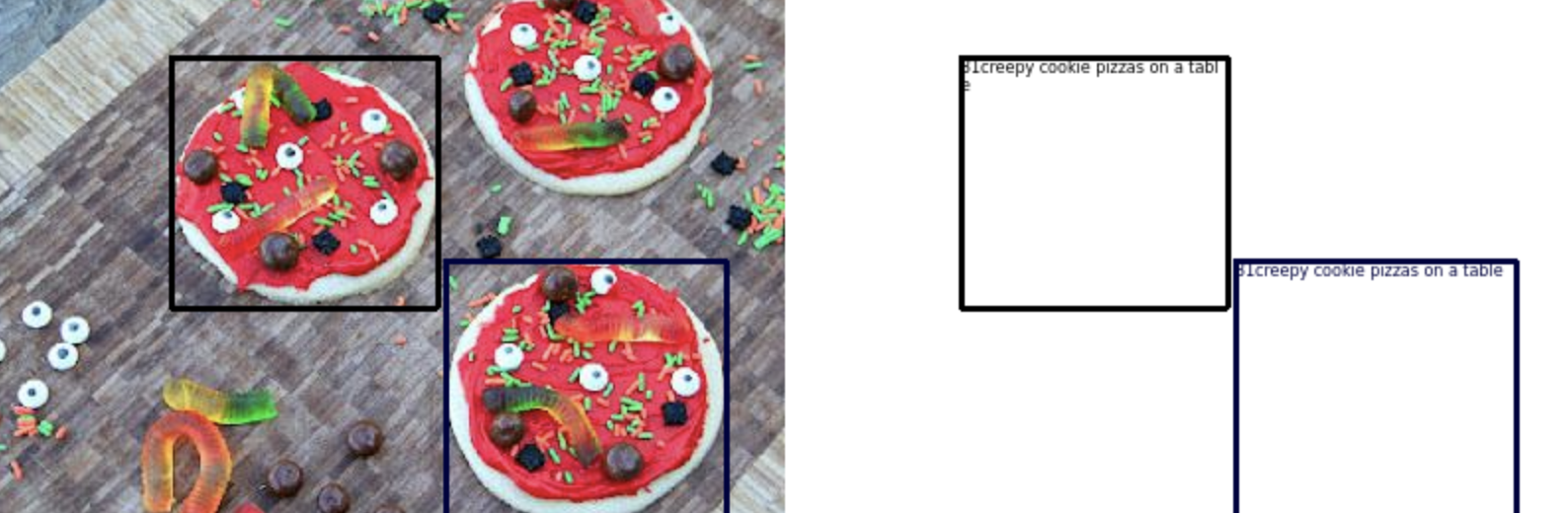}
\includegraphics[width= 0.8\textwidth]{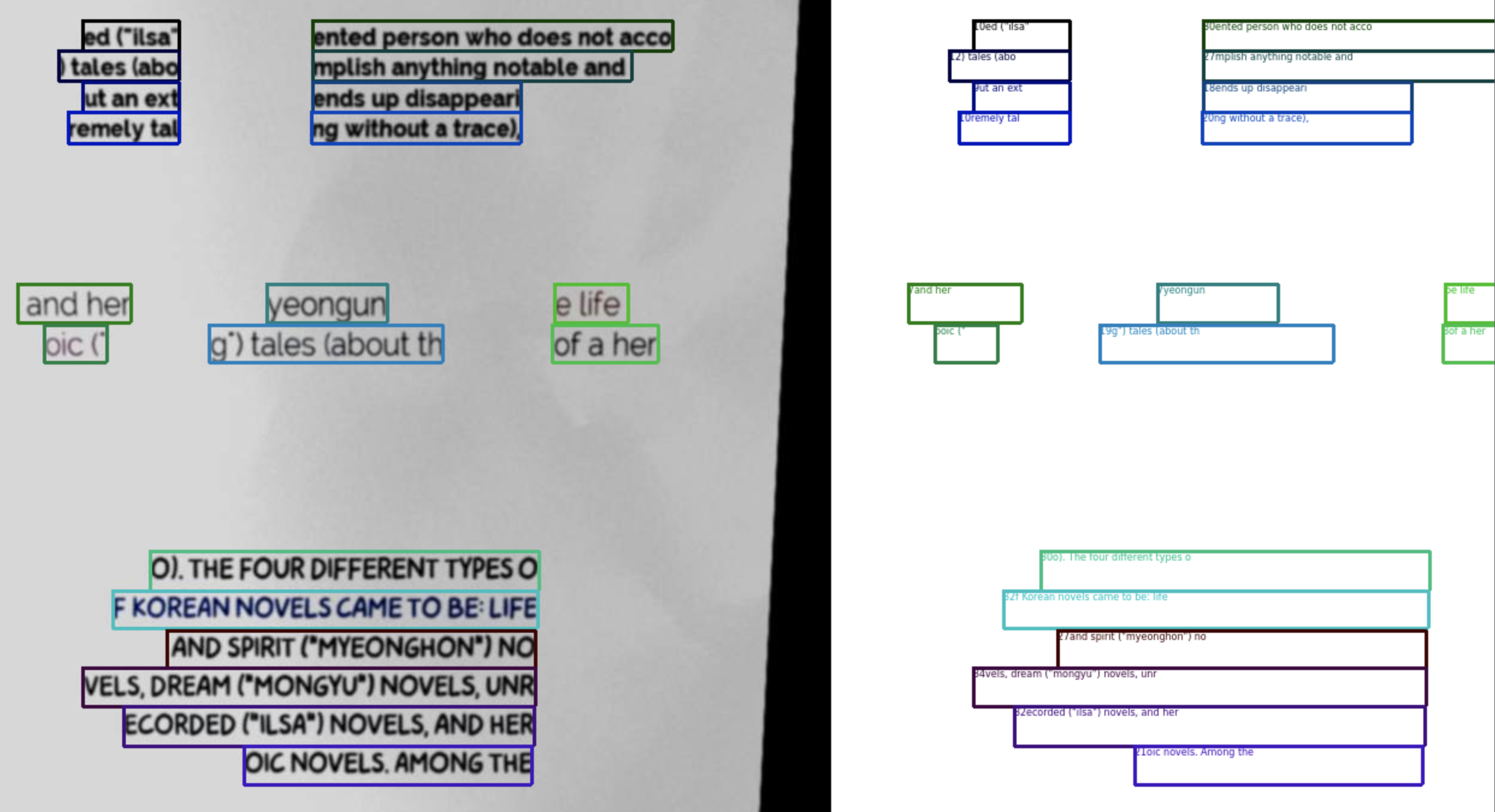}
\includegraphics[width= 1\textwidth]{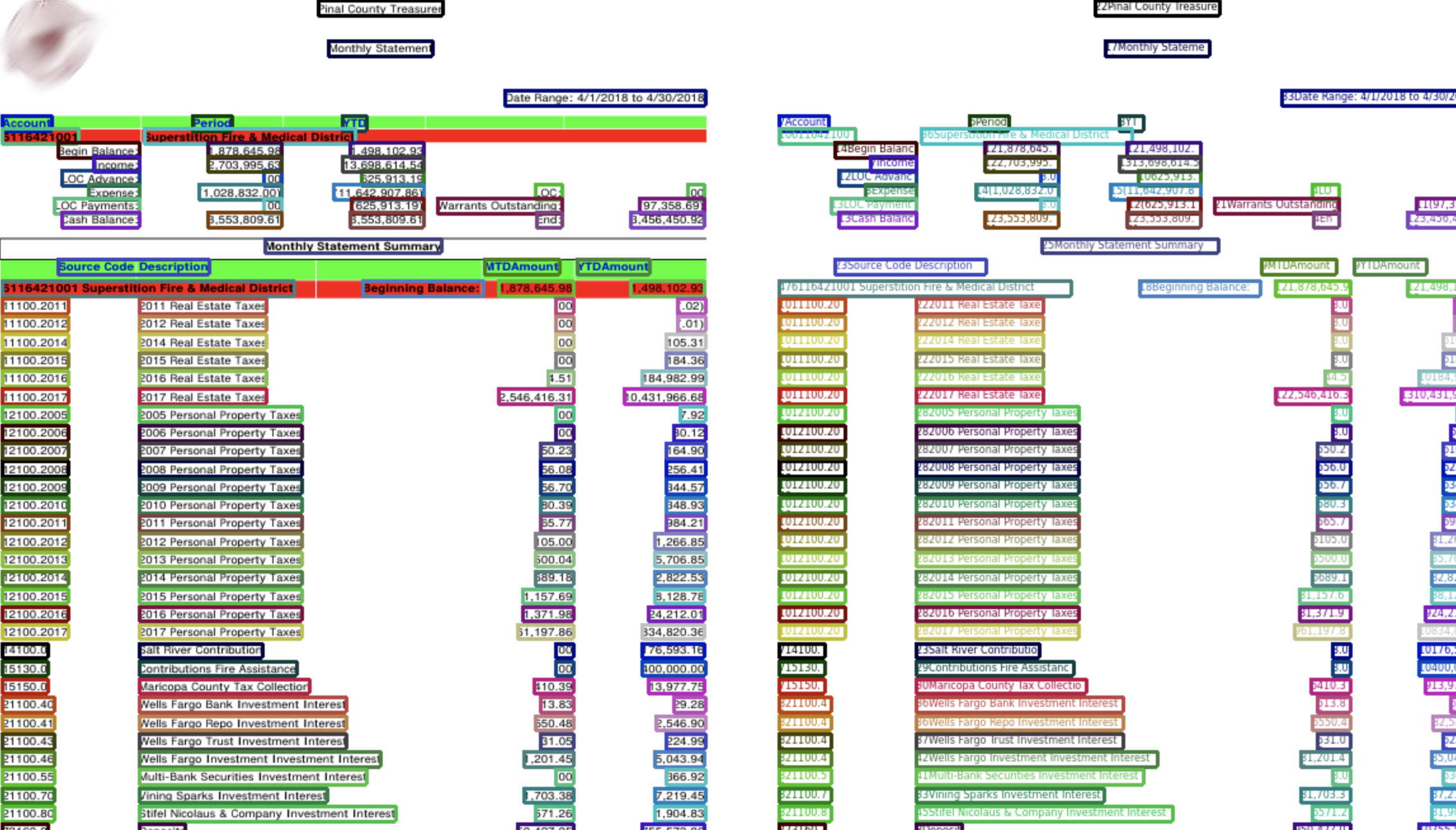}
   \caption{Visualization of the Grounding and OCR data used for training Qwen-VL}
\label{ocr_vis}
\end{figure*}

We generated the synthetic OCR dataset using Synthdog~\citep{synthdog}. Specifically, we use the COCO~\citep{lin2014microsoft} train2017 and unlabeld2017 dataset split as the natural scenery background. Then we selected 41 English fonts and 11 Chinese fonts to generate text. We use the default hyperparameters as in Synthdog. We track the generated text locations in the image and convert them to quadrilateral coordinates and we also use these coordinates as training labels. The visualization example is illustrated in the second row of Fig~\ref{ocr_vis}.

For all the PDF data we collected, we follow the steps below to pre-process the data using PyMuPDF~\citep{pymupdf} to get the rendering results of each page in a PDF file as well as all the text annotations with their bounding boxes.
\begin{enumerate}
    \item Extracting all texts and their bounding boxes for each page.
    \item Rendering each page and save them as an image file.
    \item Removing too small image.
    \item Removing images with too many or too few characters.
    \item Removing images containing Unicode characters in the ``Latin Extended-A'' and ``Latin Extended-B'' blocks.
    \item Removing images containing Unicode characters in the ``Private Use Area (PUA)'' block.
\end{enumerate}

For all HTML web pages we collected, we pre-process them in a similar approach to all the PDF data we collected, but we use Puppeteer~\citep{puppeteer} instead of PyMuPDF to render these HTML pages and get the ground truth annotation. We follow the steps below to pre-process the data.
\begin{enumerate}
    \item Extracting all texts for each webpage.
    \item Rendering each page and save them as an image file.
    \item Removing too small image.
    \item Removing images with too many or too few characters.
    \item Removing images containing Unicode characters in the ``Private Use Area (PUA)'' block.
\end{enumerate}

\section{Data Format Details of Training}

\subsection{Data Format of Multi-Task Pre-training}
\label{app:data_format_stage2}

We visualize the Multi-Task Pre-training data format in Box~\ref{mt_format}. The Box contains all 7 tasks with the black-colored text as the prefix sequence without loss and blue-colored text as the ground truth labels with loss.

\begin{tcolorbox}[colback=black!5!white,colframe=black!75!black,title=Image Captioning]
$<$img$>$cc3m/01581435.jpg$<$/img$>$Generate the caption in English: \textcolor{blue}{the beautiful flowers for design.$<$eos$>$}
\tcbsubtitle{Vision Question Answering}
$<$img$>$VG\_100K\_2/1.jpg$<$/img$>$ Does the bandage have a different color than the wrist band? Answer: \textcolor{blue}{No, both the bandage and the wrist band are white.$<$eos$>$}
\tcbsubtitle{OCR VQA}
$<$img$>$ocr\_vqa/1.jpg$<$/img$>$ What is the title of this book? Answer: \textcolor{blue}{Asi Se Dice!, Volume 2: Workbook And Audio Activities (Glencoe Spanish) (Spanish Edition)}$<$eos$>$
\tcbsubtitle{Caption with Grounding}
$<$img$>$coyo700m/1.jpg$<$/img$>$Generate the caption in English with grounding: \textcolor{blue}{Beautiful shot of $<$ref$>$bees$<$/ref$>$$<$box$>$(661,612),(833,812)$<$/box$>$$<$box$>$(120,555),(265,770) $<$/box$>$ gathering nectars from $<$ref$>$an apricot flower$<$/ref$>$$<$box$>$(224,13),(399,313) $<$/box$>$$<$eos$>$}
\tcbsubtitle{Referring Grounding}
$<$img$>$VG\_100K\_2/3.jpg$<$/img$>$$<$ref$>$the ear on a giraffe$<$/ref$>$\textcolor{blue}{$<$box$>$(176,106),(232,160) $<$/box$>$$<$eos$>$}
\tcbsubtitle{Grounded Captioning}
$<$img$>$VG\_100K\_2/4.jpg$<$/img$>$$<$ref$>$This$<$/ref$>$$<$box$>$(360,542),(476,705)$<$/box$>$ is \textcolor{blue}{Yellow cross country ski racing gloves$<$eos$>$}
\tcbsubtitle{OCR}
$<$img$>$synthdog/1.jpg$<$/img$>$OCR with grounding: \textcolor{blue}{$<$ref$>$It is managed$<$/ref$>$ $<$quad$>$ (568,121), (625,131), (624,182), (567,172)$<$/quad$>$...$<$eos$>$}
\label{mt_format}
\end{tcolorbox}

\subsection{Data Format of Supervised Fine-tuning}
\label{app:data_format_stage3}

To better accommodate multi-image dialogue and multiple image inputs, we add the string "Picture $id$:" before different images, where the $id$ corresponds to the order of image input dialogue. In terms of dialogue format, we construct our instruction tuning dataset using the ChatML~\citep{chatml} format, where each interaction's statement is marked with two special tokens ($<$im\_start$>$ and $<$im\_end$>$) to facilitate dialogue termination.

\begin{tcolorbox}[colback=black!5!white,colframe=black!75!black,title=The Dataset Format Example of ChatML]
\textcolor{blue}{$<$im\_start$>$}user

Picture 1: $<$img$>$vg/VG\_100K\_2/649.jpg$<$/img$>$What is the sign in the picture?\textcolor{blue}{$<$im\_end$>$}

\textcolor{blue}{$<$im\_start$>$}assistant

\textcolor{blue}{The sign is a road closure with an orange rhombus.$<$im\_end$>$}

\textcolor{blue}{$<$im\_start$>$}user

How is the weather in the picture?\textcolor{blue}{$<$im\_end$>$}

\textcolor{blue}{$<$im\_start$>$}assistant

\textcolor{blue}{The shape of the road closure sign is an orange rhombus.$<$im\_end$>$}
\end{tcolorbox}

During training, we ensure the consistency between prediction and training distributions by only supervising answers and special tokens (blue in the example), and not supervising role names or question prompts.  

\section{Hyperparameters}
\label{app:hyperparam}
We report the detailed training hyperparameter settings of Qwen-VL in Table~\ref{tab:hyperparam}.

\begin{table}[htbp]
    \centering
    \tablestyle{7pt}{1.3}
    \caption{Training hyperparameters of Qwen-VL}
    \begin{tabular}{l ccc}
         \toprule
         Configuration            & Pre-training & Multi-task Pre-training & Supervised Fine-tuning \\
         \midrule
         ViT init.                & Open-CLIP-bigG & Qwen-VL 1st-stage & Qwen-VL 2nd-stage \\
         LLM init.                & Qwen-7B & Qwen-7B & Qwen-VL 2nd-stage \\
         VL Adapter init.         & random & Qwen-VL 1st-stage & Qwen-VL 2nd-stage \\
         Image resolution         & $224^2$ & $448^2$ & $448^2$ \\
         ViT sequence length      & 256 & 1024 & 1024 \\
         LLM sequence length      & 512 & 2048 & 2048\\
         Learnable query numbers  & 256 & 256 & 256\\
         Optimizer                & \multicolumn{3}{c}{AdamW} \\
         Optimizer hyperparameter & \multicolumn{3}{c}{$\beta_{1}=0.9, \beta_{2}=0.98, eps=1e^{-6}$} \\
         Peak learning rate       & $2e^{-4}$ & $5e^{-5}$ & $1e^{-5}$ \\
         Minimum learning rate    & $1e^{-6}$ & $1e^{-5}$ & $1e^{-6}$ \\
         ViT learning rate decay  & 0.95 & 0.95 & 0 \\
         ViT Drop path rate       & \multicolumn{3}{c}{0} \\
         Learning rate schedule   & \multicolumn{3}{c}{cosine decay} \\
         Weight decay             & \multicolumn{3}{c}{0.05} \\
         Gradient clip            & \multicolumn{3}{c}{1.0} \\
         Training steps           & 50k & 19k & 8k \\
         Warm-up steps            & 500 & 400 & 3k \\
         Global batch size        & 30720 & 4096 & 128 \\
         Gradient Acc.            & 6 & 8 & 8 \\
         Numerical precision      & \multicolumn{3}{c}{$\mathtt{bfloat16}$} \\
         Optimizer sharding       & \multicolumn{3}{c}{\ding{51}} \\
         Activation checkpointing & \multicolumn{3}{c}{\ding{55}} \\
         Model parallelism        & \ding{55} & 2 & 2 \\
         Pipeline parallelism     & \multicolumn{3}{c}{\ding{55}} \\
         \bottomrule
    \end{tabular}
    \label{tab:hyperparam}
\end{table}

In the first pre-training stage, the model is trained using AdamW optimizer with $\beta_{1}=0.9, \beta_{2}=0.98, eps=1e^{-6}$. We use the cosine learning rate schedule and set the maximum learning rate of $2e^{-4}$ and minimum of $1e^{-6}$ with a linear warm-up of 500 steps. We use a weight decay of $5e^{-2}$ and a gradient clipping of $1.0$. For the ViT image encoder, we apply a layer-wise learning rate decay strategy with a decay factor of $0.95$. 
The training process uses a batch size of 30720 for the image-text pairs, and the entire first stage of pre-training lasts for 50,000 steps, consuming approximately 1.5 billion image-text samples and 500 billion image-text tokens.

In the second multi-task training stage, we increase the input resolution of the visual encoder from $224\times 224$ to $448\times 448$, reducing the information loss caused by image down-sampling. We unlocked the large language model and trained the whole model. The training objective is the same as the pre-training stage. We use AdamW optimizer with $\beta_{1}=0.9, \beta_{2}=0.98, eps=1e^{-6}$. We trained for 19000 steps with 400 warm-up steps and a cosine learning rate schedule. Specifically, we use the model parallelism techniques for ViT and LLM.

\section{Summary of the evaluation benchmarks}
\label{app:benchmark}

We provide a detailed summary of the used evaluation benchmarks and corresponding metrics in Table~\ref{tab:benchmark}.

\begin{table}[ht]
    \centering
    \caption{Summary of the evaluation benchmarks.}
    \scripttablestyle{3pt}{1.05}
    \begin{tabular}{l|l|l|l|l}
        \toprule
         Task & Dataset & Description & Split & Metric  \\
         \midrule
         \multirow{2}{*}{Image Caption} & Nocaps & Captioning of natural images & val & CIDEr($\uparrow$) \\
         & Flickr30K & Captioning of natural images & karpathy-test & CIDEr($\uparrow$) \\
         \midrule
         \multirow{5}{*}{General VQA} & VQAv2 & VQA on natural images & test-dev & VQA Score($\uparrow$) \\
         & OKVQA & VQA on natural images requiring outside knowledge & val & VQA Score($\uparrow$) \\
         & GQA & VQA on scene understanding and reasoning & test-balanced & EM($\uparrow$) \\
         & ScienceQA-Img & Multi-choice VQA on a diverse set of science topics & test & Accuracy($\uparrow$) \\
         & VizWiz & VQA on photos taken by people who are blind & test-dev & VQA Score($\uparrow$)\\
         \midrule
         \multirow{5}{*}{Text-oriented VQA} & TextVQA & VQA on natural images containing text & val & VQA Score($\uparrow$) \\
         & DocVQA & VQA on images of scanned documents & test & ANLS($\uparrow$) \\
         & ChartQA & VQA on images of charts & test & Relaxed EM($\uparrow$)\\
         & OCRVQA & VQA on images of book covers & test & EM($\uparrow$) \\
         & AI2Diagram & VQA on images of scientific diagrams & test & EM($\uparrow$) \\
         \midrule
         & RefCOCO & Refer grounding on natural images & val \& testA \& testB & Accuracy($\uparrow$) \\
         Refer Expression & RefCOCO+ & Refer grounding on natural images & val \& testA \& testB & Accuracy($\uparrow$) \\
         Comprehension & RefCOCOg & Refer grounding on natural images & val \& test & Accuracy($\uparrow$) \\
         & GRiT & Refer grounding on natural images & test & Accuracy($\uparrow$) \\
         \midrule
         \multirow{3}{*}{Instruction Following} & TouchStone & Open-ended VL instruction following benchmark & English \& Chinese & GPT-$4$ Score ($\uparrow$) \\
          & MME & Open-ended VL Benchmark by yes/no questions & Perception \& Cognition & Accuracy ($\uparrow$) \\
          & Seed-Bench & Open-ended VL Benchmark by Multi-choice VQA & Image \& Video & Accuracy ($\uparrow$) \\
         \bottomrule
    \end{tabular}
    \label{tab:benchmark}
\end{table}

\section{Additional experimental details}

\subsection{Convergence of the Pre-training Stage}
\label{app:first_stage}

\begin{figure*}[ht]
\centering
\includegraphics[width= 1\textwidth]{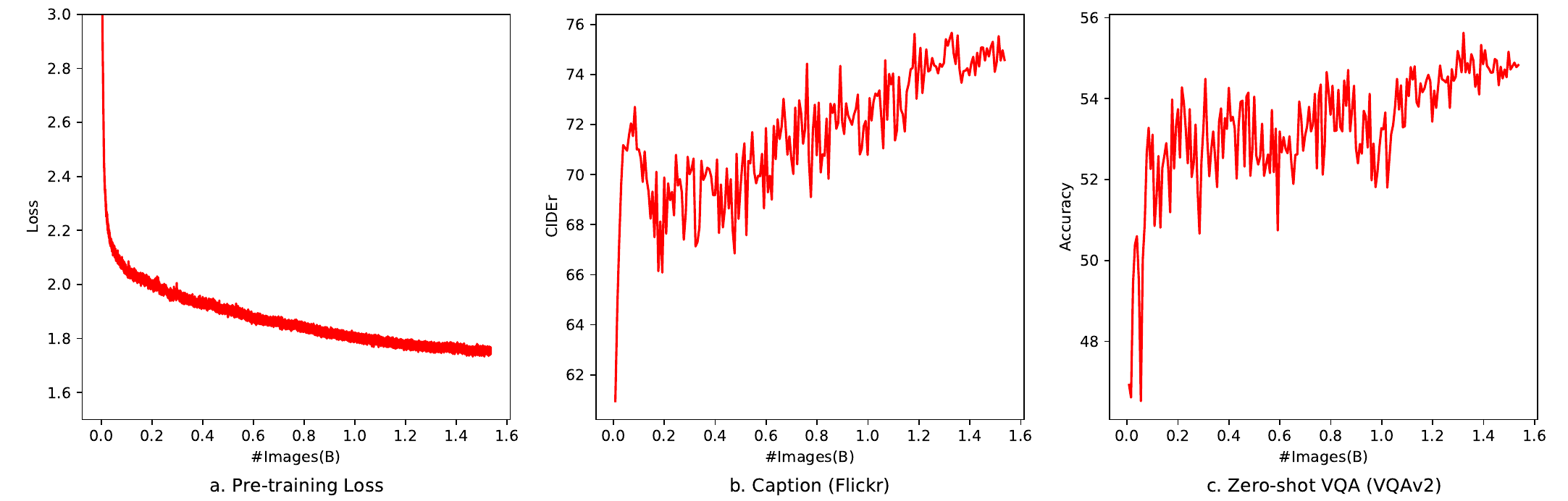}
   \caption{Visualization of the Convergence of the Pre-training Stage}
\label{fig:stage1}
\end{figure*}

In Figure \ref{fig:stage1}, we show the convergence of the Pre-training Stage (stage one). The whole models are trained using BFloat16
mixed precision, the batch size is 30720, and the learning rate is $2e^{-4}$. All images are only trained once (one epoch). The training loss decreases steadily with the increase of the number of training pictures. Note that, the pre-training stage (Stage one) has no VQA data being added, but the Zero-shot VQA score increases amidst fluctuations.

\subsection{Number of Learnable Queries in the Vision-Language Adapter}
\label{app:n_queries}

The vision-language adapter uses cross-attention to compress the visual feature sequence by a set of learning queries of length. Too few queries can lead to the loss of some visual information, while too many queries may reduce in greater convergence difficulty and computational cost.

\begin{figure*}[ht]
\centering
\includegraphics[width= 0.85\textwidth]{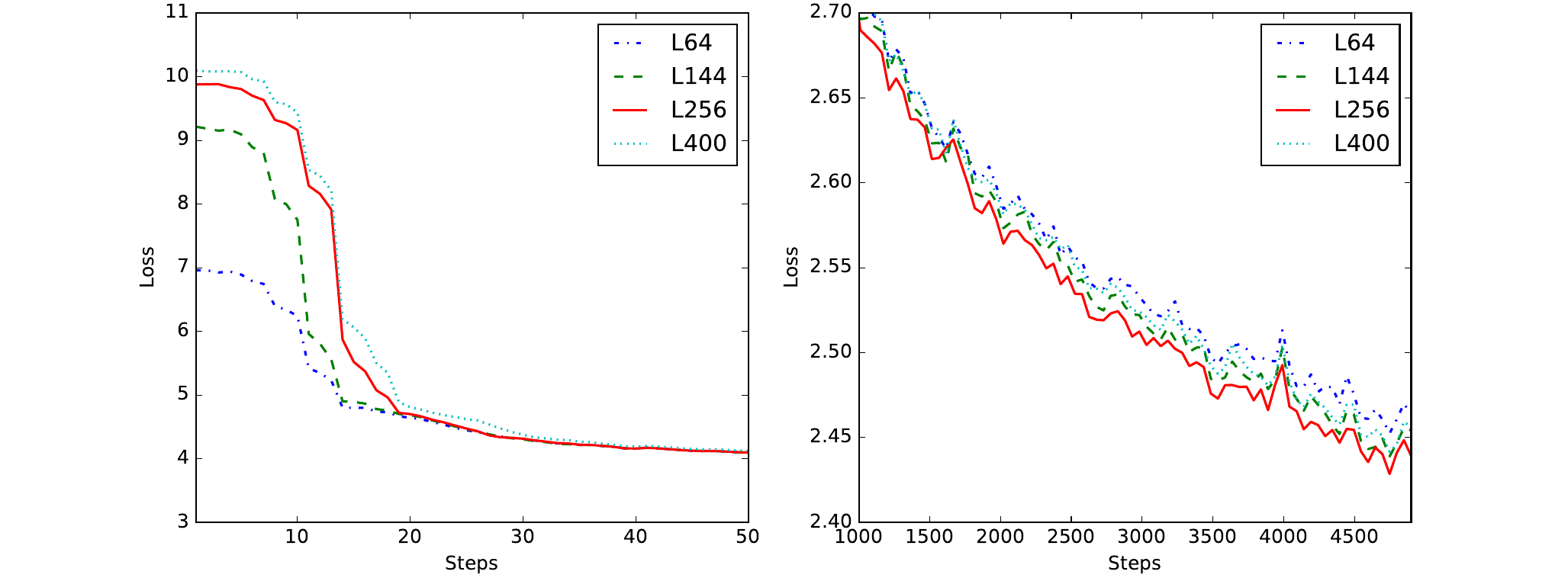}
   \caption{Visualization of the training loss when using different compressed feature lengths of the vision-language adapter. The left depicts the initial training loss (within 50 steps), and the right depicts the loss in convergence (1k-5k steps). In the legend, L64 denotes that the adapter uses 64 queries to compress the visual feature sequence to a fixed length of 64, and so on. The loss curves have been smoothed to avoid shading owing to fluctuations.}
\label{fig:ablation_adapter}
\end{figure*}

An ablation experiment is conducted on the number of learnable queries in the vision-language adapter. We used ViT-L/14 as the visual encoder and the $224\times224$ resolution picture as input, so the sequence length of ViT's output is $(224/14)^2=256$. As shown in the left part of Figure \ref{fig:ablation_adapter}, the fewer queries used at the beginning of training, the lower the initial loss. However, with convergence, too many or too few queries will cause convergence to slow down, as shown in the right part of Figure \ref{fig:ablation_adapter}. Considering that the second training stage (Multi-task Pre-train) applies 448*448 resolution, where the sequence length of ViT's output is $(448/14)^2=1024$. Too few queries can result in more information being lost. We finally chose to use 256 queries for the vision-language adapter in Qwen-VL.

\subsection{Window Attention vs Global Attention for Vision Transformer}
\label{app:window_attention}

Using a high-resolution Vision Transformer in the model will significantly increase the computational cost. One possible solution to reduce the computational cost of the model is to use Window Attention in the Vision Transformer, i.e., to perform Attention only in a window of $224 \times 224$ in most layers of the ViT part of the model, and to perform Attention for the full $448 \times 448$ or $896 \times 896$ image in a small number of layers (e.g. 1 out of every 4 layers) of the ViT part of the model.

To this end, we conducted ablation experiments to compare the performance of the model when using Global Attention and Window Attention for ViT. We compare the experimental results for analysing the trade-off between computational efficiency and convergence of the model.

\begin{figure*}[ht]
\centering
\includegraphics[width= 1\textwidth]{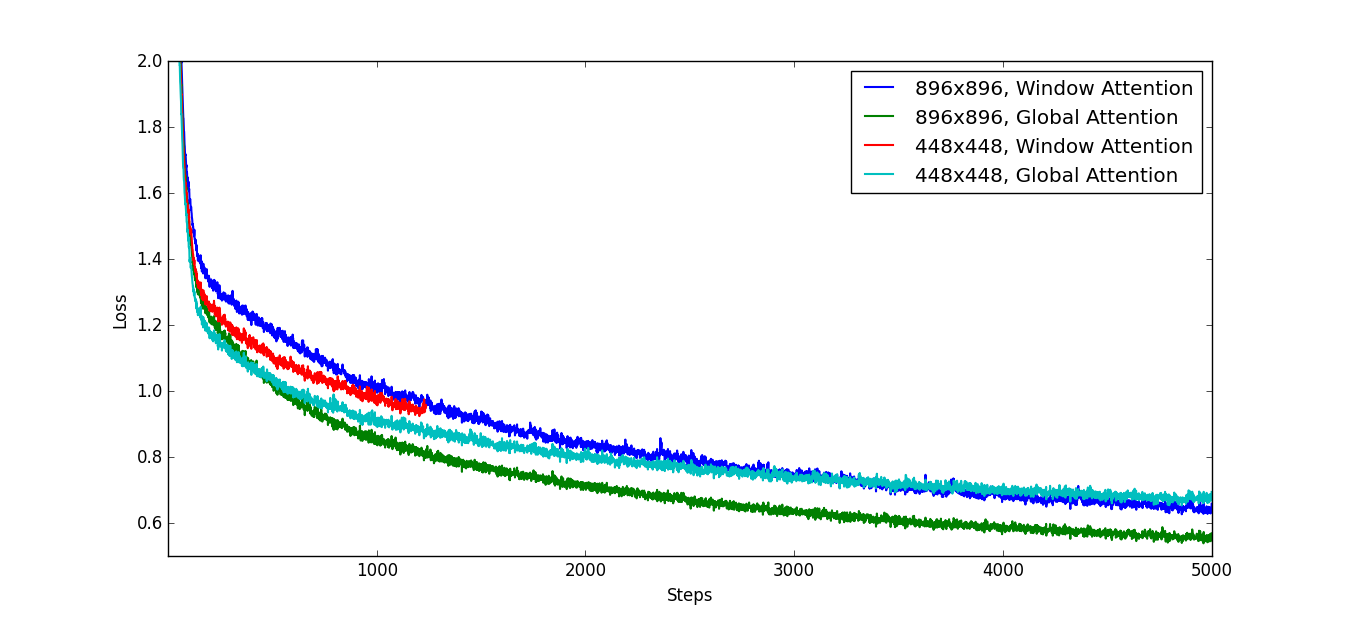}
   \caption{Visualization of the Loss when using Window Attention vs Global Attention}
\label{fig:ablation_attention}
\end{figure*}

\begin{table}[htbp]
    \centering
    \tablestyle{7pt}{1.3}
    \caption{Training speed of Window Attention vs Global Attention for different input image resolutions}
    \begin{tabular}{l c}
         \toprule
         Model input resolution \& Attention type            & Training speed  \\
         \midrule
         $448\times448$, Global Attention        & ~10s / iter \\
         $448\times448$, Window Attention            & ~9s / iter \\
         $896\times896$, Global Attention      & ~60s / iter \\
         $896\times896$, Window Attention       & ~25s / iter \\
         \bottomrule
    \end{tabular}
    \label{tab:ablation_attention}
\end{table}

As shown in Figure \ref{fig:ablation_attention} and Table \ref{tab:ablation_attention}, the loss of the model is significantly higher when Window Attention instead of Vanilla Attention is used. And the training speeds for both of them are similar. Therefore, we decided to use Vanilla Attention instead of Window Attention for the Vision Transformer when training Qwen-VL.

The reason we don't use Window Attention with $896\times896$ resolution is that its training speed is too slow for us. Although it reaches a loss value similar to model with $448 \times 448$ resolution input at 5000 steps. It takes almost 2.5 times longer to train than the model with $448 \times 448$ resolution input.

\subsection{Performance on Pure-text Tasks}

In order to study the effect of multi-modal training on pure-text ability, we show the performance of pure-text tasks of Qwen-VL compared to open-source LLM in Table \ref{tab:pure_text}.

Qwen-VL uses an intermediate checkpoint of Qwen-7B as the LLM initialization. The reason why we did not use the final released checkpoint of Qwen-7B is that Qwen-VL and Qwen-7B were developed at a very similar period. Because Qwen-VL has a good initialization on LLM by Qwen-7B, it is comparable to many text-only LLMs on pure-text tasks.

\begin{table}[ht]
\centering
\caption{Performance on Pure-text Benchmarks of Qwen-VL compared to open-source LLM. Due to the introduction of pure-text data in the multi-task training and SFT stage, Qwen-VL do not compromise any pure-text ability.}
\tablestyle{7pt}{1.3}
\begin{tabular}{@{}l|ccc@{}}
\toprule
Model & MMLU & CMMLU & C-Eval \\ \midrule
LLaMA-7B & 35.1 & 26.8 & - \\
LLaMA2-7B & 46.8 & 31.8 & 32.5 \\
Baichuan-7B & 42.3 & 44.4 & 42.8 \\
Baichuan2-7B & 54.2 & 57.1 & 54.0 \\
ChatGLM2-6B & 47.9 & 48.8 & 51.7 \\
InternLM-7B & 51.0 & 51.8 & 52.8 \\ 
Qwen-7B (final released) & 58.2 & 62.2 & 63.5 \\ \midrule
Qwen-7B (intermediate, use as Qwen-VL's LLM initialization) & 49.9 & - & 48.5 \\
Qwen-VL & 50.7 & 49.5 & 51.1 \\ \bottomrule
\end{tabular}
\label{tab:pure_text}
\end{table}

Furthermore, in the multi-task training and SFT stages, Qwen-VL not only utilizes visual and language-related data but also incorporates pure-text data for training. The purpose of this is to prevent the catastrophic forgetting of text comprehension by leveraging the information from pure-text data. The results in Table \ref{tab:pure_text} indicate that the Qwen-VL model does not exhibit any degradation in terms of its pure text capability and even demonstrates improvement after multi-task training.

\end{document}